%% file: main.tex
\documentclass[14pt]{article}

\usepackage{arxiv2}

\usepackage[utf8]{inputenc} 
\usepackage[T1]{fontenc}    
\usepackage{hyperref}       
\usepackage{url}            
\usepackage{booktabs}       
\usepackage{amsfonts}       
\usepackage{nicefrac}       
\usepackage{microtype}      
\usepackage{lipsum}
\usepackage{graphicx}
\graphicspath{ {./images/} }

\usepackage[fontsize=12pt]{scrextend}
\usepackage{wrapfig}
\usepackage[linesnumbered,ruled,vlined]{algorithm2e}
\def\code#1{\texttt{#1}}
\usepackage{footmisc}
\usepackage{amsthm,bm,amsfonts,amsmath,xcolor}
\newtheorem{prop}{Proposition}

\usepackage[numbers]{natbib}

\usepackage[normalem]{ulem}
\newcommand\remove{\bgroup\markoverwith{\textcolor{orange}{\rule[.5ex]{2pt}{1pt}}}\ULon}

\newcommand{\removelatexerror}{\let\@latex@error\@gobble}
 
\newcommand{\x}{\mathbf{x}}
\newcommand{\X}{\mathbf{X}}

\renewcommand{\P}{\mathbb{P}}
\newcommand{\I}{\mathbb{I}}
\newcommand{\E}{\mathbb{E}}

\usepackage{tabularx}
\usepackage{bm}
\usepackage{multicol} 
\usepackage{multirow}
\usepackage{threeparttable}
\usepackage{float}
\restylefloat{table}

\title{A unified
framework for dataset shift diagnostics}

\author{
 Felipe Maia Polo \\
  Department of Statistics\\
  University of Michigan\\
  United States of America \\
   \And
  Rafael Izbicki \\
  Department of Statistics\\
  Federal University of São Carlos\\
  Brazil \\
  \And
  Evanildo Gomes Lacerda Jr \\
  Trustly\\
  Brazil \\
  \AND
  Juan Pablo Ibieta-Jimenez \\
  Experian LatAm DataLab \\
  Brazil \\
  \And
  Renato Vicente \\
  Department of Applied Mathematics\\
  Institute of Mathematics and Statistics \\
  University of São Paulo \\
  Brazil \\
}

\begin{document}
\maketitle
\begin{abstract}
Supervised learning techniques typically assume training data originates from the target population. Yet, in reality, \emph{dataset shift} frequently arises, which, if not adequately taken into account, may decrease the performance of their predictors. In this work, we propose a novel and flexible framework called \textit{DetectShift}\footnote{Our implementation for \textit{DetectShift} can be found in \url{https://github.com/felipemaiapolo/detectshift}.} that quantifies and tests for multiple dataset shifts, encompassing shifts in the distributions of $(\X, Y)$, $\X$, $Y$, $\X|Y$, and $Y|\X$. \textit{DetectShift} equips practitioners with insights into data shifts, facilitating the adaptation or retraining of predictors using both source and target data. This proves extremely valuable when labeled samples in the target domain are limited. The framework utilizes test statistics with the same nature to quantify the magnitude of the various shifts, making results more interpretable. It is versatile, suitable for regression and classification tasks, and accommodates diverse data forms—tabular, text, or image. Experimental results demonstrate the effectiveness of \textit{DetectShift} in detecting dataset shifts even in higher dimensions\footnote{Corresponding author: Felipe Maia Polo <felipemaiapolo@gmail.com>.}. 
\end{abstract}


\input{intro}

\input{method}

\input{exp}

\input{conclusions}

\newpage
\bibliographystyle{plainnat}
\bibliography{refs2}

\newpage
\appendix
\input{append.tex}

\end{document}

%% file: intro.tex
\section{Introduction}

In machine learning applications, it is conventionally assumed that training data originates from the distribution of interest. We say a dataset shift has happened when that assumption does not hold. Formally, we have dataset shift when the joint distribution of features ($\X$) and labels ($Y$) associated with the training sample, also known as the source distribution, $P^{(1)}_{\X,Y}$, and the distribution of interest, also known as the target distribution, $P^{(2)}_{\X,Y}$, are different. Real-world applications, spanning fields like finance \cite{speakman2018three}, health \cite{finlayson2021clinician}, technology \cite{li2010application}, and physics \cite{freeman2017unified}, often grapple with the challenges posed by dataset shifts. Unfortunately, dataset shift may substantially decrease the predictive power of machine learning models if it is not adequately addressed \cite{sugiyama2012machine}.

When labeled data from the target distribution is scarce, addressing dataset shifts typically necessitates assumptions about the relationship between the source and target distributions. Different assumptions are translated into different types of shift \cite{moreno2012unifying}, and every kind of shift demands specific adaptation methods \cite{saerens2002adjusting,sugiyama2012machine, maity2021linear}. For instance, if $P^{(1)}_{\X}\neq P^{(2)}_{\X}$ but $P^{(1)}_{Y|\X}= P^{(2)}_{Y|\X}$, dataset shift adaptation can be performed by using importance weighting on the training data \cite{sugiyama2012machine,izbicki2017photo,maia2022effective}.
Similarly, if $P^{(1)}_{Y}\neq P^{(2)}_{Y}$ but $P^{(1)}_{\X|Y}= P^{(2)}_{\X|Y}$, dataset shift adaptation can be performed by importance weighting \cite{lipton2018detecting} or re-calibrating posterior probabilities via Bayes' theorem \cite{saerens2002adjusting,vaz2019quantification}. Therefore, to successfully adapt prediction algorithms in a dataset shift setting, practitioners need to know not only \emph{if} dataset shift occurs but also \emph{which type} of shift happens for the data at hand.

This study introduces \textit{DetectShift}, a novel unified framework capable of accurately quantifying and testing for diverse types of dataset shifts. This framework provides valuable insights to practitioners on optimally handling changes in the data, particularly in scenarios where labeled samples from the target domain are scarce. By using our framework, practitioners can detect specific types of shifts. Consequently, they can use that information to use both source and target data to retrain or adapt their predictors, which is especially appealing in situations where naive model retraining using only data points from $P^{(2)}_{\X,Y}$ is not feasible\footnote{See \cite{saerens2002adjusting,lipton2018detecting,sugiyama2012machine, maity2021linear} for some examples on how to retrain or adapt predictors using not only target samples. \label{fn:retrain}}. In Section \ref{sec:exp}, we show in practice one example of how the insights provided by \textit{DetectShift} can help practitioners in a prediction task in which target-labeled samples are not used to retrain the classifier.

The remainder of this paper is organized as follows. Firstly, in Section \ref{sec:related}, we provide an overview of recent works relevant to our research, highlighting our contributions to the existing literature. Moving on to Section \ref{sec:method}, we present our framework called \textit{DetectShift}, explaining its conceptual basis and rationale. To demonstrate the effectiveness of our approach across various scenarios, we apply \textit{DetectShift} to both artificial and real data in Section \ref{sec:exp}. Finally, in the concluding sections, we thoroughly discuss our findings, contributions, limitations, and potential avenues for further exploration.

\subsection{Contributions}

The main contributions of \textit{DetectShift} are that:

\begin{itemize}
    \item It facilitates quantification and formal testing of shifts in the distributions of $(\X, Y)$, $\X$, $Y$, $\X|Y$, and $Y|\X$. Identifying different types of shifts helps practitioners gain insights about changes in their data. Those insights allow them to leverage source and target data to retrain or adapt their predictors\footref{fn:retrain}, which is especially beneficial when target labels are scarce.
    
    \item It utilizes test statistics with the same nature to quantify each type of dataset shift, enabling practitioners to compare the magnitude of different shifts and gain meaningful insights. Furthermore, empirical evidence demonstrates that these test statistics result in powerful shift detectors even in higher dimensions.
    
    \item It can be applied to regression and classification tasks. Additionally, it can be employed in conjunction with virtually any type of data, including tabular, text, and image data.
\end{itemize}

\section{Related work}\label{sec:related}

Dataset shift detection methods can have different objectives. For instance, specific approaches monitor the test error of a predictor over time, triggering an alarm when there is a significant degradation in its performance \cite{gama2004learning,yu2019concept, podkopaev2021tracking}. Other approaches are specifically designed to provide a deeper understanding of how data distribution has changed by detecting specific types of shifts, e.g., changes in the distribution of features. {Our research aligns closely with the latter category, which we will delve into next.

Source and target distributions can be different in many ways. For instance, if $P^{(1)}_{Y}\neq P^{(2)}_{Y}$ but $P^{(1)}_{\X|Y}= P^{(2)}_{\X|Y}$, then the dataset shift can be characterized as label shift. If labels from the target domain are unavailable, label shift detection and quantification are not straightforward but can be accomplished under some conditions \cite{lipton2018detecting,vaz2019quantification}. The method proposed by \citet{lipton2018detecting}, for example, exploits arbitrary black box classifiers and their confusion matrices to estimate $P^{(2)}_{Y}$. In a different direction, various methods are available when we are interested in detecting or understanding changes in the marginal distribution of possibly high-dimensional features $\X$. Detecting shifts in $P_{\X}$ can be solved using two-sample tests \cite{rabanser2019failing, lopez2016revisiting}, and some extensions are possible. For example, \citet{jang2022sequential} proposes an online approach in which data is seen sequentially, and \citet{wijaya2021failing} proposes extracting concept-based features from images (e.g., rotation, shape) and then testing for shifts of those concepts. Moreover, \citet{ginsberg2023a} proposes a method for \textit{harmful} covariate shift detection while \citet{luo2022online} proposes a martingale-based approach in which false alarms are controlled over time. For objectives centered on detecting shifts in conditional distributions like $P_{Y|\X}$, several methods emerge. For instance, \citet{schrouff2022diagnosing} proposes using a conditional independence test to detect shifts in conditional distributions in the context of algorithmic fairness. In contrast, \citet{vovk2020testing} proposes a martingale-based approach to detect online changes in $P_{\X|Y}$ considering classification problems.

While identifying specific types of shifts is essential, it is typically insufficient for making informed decisions. Ideally, practitioners should access a unified framework that evaluates multiple dataset shifts, guiding them on optimal next steps. That framework would also allow the practitioner to compare the strength of the several shift types, permitting a more interpretable analysis. \citet{webb2018analyzing} proposes a framework with those characteristics. The authors use the total variation (TV) distance between two probability distributions to measure shifts in $P_{Y}$, $P_{\X}$, $P_{Y|\X}$, and $P_{\X|Y}$, to provide the practitioner with valuable insights. Their approach is limited, however. Estimating the TV distance between two probability distributions is challenging when variables are not discrete or their dimensionality is high (see Figure \ref{fig:power2} in our experiments section).
Moreover, \citet{webb2018analyzing} does not suggest using formal hypothesis testing. Thus, false alarm control is not guaranteed. Building upon \citet{webb2018analyzing}, our work augments their framework, offering adaptability to high-dimensional/continuous data and integrating rigorous hypothesis testing.

%% file: method.tex
\section{Methodology}\label{sec:method}

We observe two datasets,  $\mathcal{D}^{(1)}$
and $\mathcal{D}^{(2)}$, where  $$\mathcal{D}^{(i)}=\left\{\left(\X^{(i)}_1,Y^{(i)}_1\right),\ldots,\left(\X^{(i)}_{n^{(i)}},Y^{(i)}_{n^{(i)}}\right)\right\},$$ 
for $i=1,2$.
We assume that  all observations from the same dataset are i.i.d. and that the datasets are
independent of each other. 
We denote by $P^{(i)}_{\X,Y}$ the distribution associated with an observation from the $i$-th dataset, where $i=1$ stands for source and $i=2$ for the target. We aim to quantify and test which types of dataset shifts occur from source to target domain. The null hypotheses we want to test are 
\begin{itemize}
\addtolength\itemsep{.2mm}
    \item \textbf{[Total Dataset Shift]}  
$H_{0,\text{D}}:P^{(1)}_{\X,Y}=P^{(2)}_{\X,Y}~~$ 
      \item \textbf{[Feature Shift]} 
    $H_{0,\text{F}}:P^{(1)}_{\X}=P^{(2)}_{\X}~~$
    \item  \textbf{[Response Shift]} 
    $H_{0,\text{R}}:P^{(1)}_{Y}=P^{(2)}_{Y}~~$ 
    \item \textbf{[Conditional Shift - Type 1]}  
$H_{0,\text{C1}}:P^{(1)}_{\X|Y}=P^{(2)}_{\X|Y}~~$ ($P^{(2)}_{Y}$- almost surely)
    \item \textbf{[Conditional Shift - Type 2}] 
    $H_{0,\text{C2}}: P^{(1)}_{Y|\X}=P^{(2)}_{Y|\X}~~$ ($P^{(2)}_{\X}$- almost surely)
\end{itemize}
Section \ref{sec:test_stat}  describes how to obtain statistics that can quantify the amount of each type of dataset shift and
Section \ref{sec:hypo_test} shows how to use the statistics to test each type of shift's occurrence formally.

\subsection{Test statistics}\label{sec:test_stat}

Our statistics are based on the Kullback-Leibler (KL) divergence \cite{kullback1951information,polyanskiy2022information}, a well-known measure to describe discrepancies between probability measures. Formally, the KL divergence between two probability distributions  $P$ and $Q$ is defined as $$\textup{KL}(P||Q):=\int \log\left(\frac{dP}{dQ} \right)dP,$$
where $\frac{dP}{dQ}$ is the Radon–Nikodym (R-N) derivative of $P$ with respect to $Q$. If both $P$ and $Q$ are (Lebesgue) continuous distributions, the R-N derivative is simply a density ratio $p/q$. For the R-N derivative to be well defined, we need $P$ to be absolutely continuous with respect to $Q$ \cite{polyanskiy2022information}, meaning that the support of $P$ is a subset of the support of $Q$. From now on, we assume that $P^{(2)}_{\X,Y}$ is absolutely continuous with respect to $P^{(1)}_{\X,Y}$. Then, we use the following quantities to measure each of the shifts described at the beginning of Section \ref{sec:method}:
\begin{itemize}
\addtolength\itemsep{.2mm}
    \item \textbf{[Total Dataset Shift]}  
    $\textup{KL}_{\X,Y}:=\textup{KL}(P^{(2)}_{\X,Y}||P^{(1)}_{\X,Y})$
    \item \textbf{[Feature Shift]}
    $\textup{KL}_{\X}:=\textup{KL}(P^{(2)}_{\X}||P^{(1)}_{\X})$
    \item  \textbf{[Response Shift]} 
    $\textup{KL}_{Y}:=\textup{KL}(P^{(2)}_{Y}||P^{(1)}_{Y})$
    \item \textbf{[Conditional Shift - Type 1]}  
    $\textup{KL}_{\X|Y}:=\mathbb{E}_{P^{(2)}_{Y}}\left[\textup{KL}(P^{(2)}_{\X|Y}||P^{(1)}_{\X|Y})\right]$
    \item \textbf{[Conditional Shift - Type 2]} 
    $\textup{KL}_{Y|\X}:=\mathbb{E}_{P^{(2)}_{\X}}\left[\textup{KL}(P^{(2)}_{Y|\X}||P^{(1)}_{Y|\X})\right]$
\end{itemize}
The following proposition states that we can rewrite the null hypotheses we want to test in terms of the quantities above.

\begin{prop}\label{prop:hyp}
The hypotheses $H_{0,\text{D}}$, $H_{0,\text{F}}$, $H_{0,\text{R}}$, $H_{0,\text{C1}}$, and $H_{0,\text{C2}}$ can be rewritten equivalently as follows
$$ H_{0,\text{D}}: \textup{KL}_{\X,Y}=0,~~~~ H_{0,\text{F}}: \textup{KL}_{\X}=0,~~~~ H_{0,\text{R}}: \textup{KL}_{Y}=0$$ $$ H_{0,\text{C1}}: \textup{KL}_{\X|Y}=0,~~~~~ H_{0,\text{C2}}: \textup{KL}_{Y|\X}=0$$
\end{prop}

The proof follows from the elementary KL divergence properties \cite{polyanskiy2022information}, e.g., $\textup{KL}(P||Q)=0$ if and only if $P=Q$. For the conditional shifts, see that $\textup{KL}(P^{(2)}_{Y|\X}||P^{(1)}_{Y|\X})$ and $\textup{KL}(P^{(2)}_{\X|Y}||P^{(1)}_{\X|Y})$ are non-negative random variables, consequently, their expected values are zero if and only if they are zero almost surely (with probability one).

To quantify and test the different types of shift, we use estimators of the parameters $\textup{KL}_{\X,Y}$, $\textup{KL}_{\X}$, $\textup{KL}_{Y}$, $\textup{KL}_{\X|Y}$, and $\textup{KL}_{Y|\X}$ as test statistics. That is a reasonable choice because (i) our null hypotheses can be equivalently written in terms of such estimable parameters, and (ii) the magnitude of the statistics is directly related to the shift intensities. These suggest that tests based on these statistics will be powerful in detecting shifts. Moreover, all parameters are integrals computed using the target distribution. Thus, they give more weight to regions of the feature/label space to which most target data points belong, letting us focus on regions that matter.

To estimate $\textup{KL}_{\X,Y}$, $\textup{KL}_{\X}$, and $\textup{KL}_{Y}$, we first use\footnote{When $Y$ is discrete, $\textup{KL}_{Y}$ can also be estimated by using a plug-in estimator described in the appendix.} training data and the probabilistic classification method for density ratio estimation \cite{sugiyama2012density,dalmasso2021likelihood}, also known as odds-trick, to estimate the Radon–Nikodym derivative between the two probability distributions. Then, we use test data to estimate the divergences.
More precisely, we first create the augmented dataset 
\begin{align*}
\mathcal{D}=\left\{\left(\X_1,Y_1,Z_1\right),\ldots,\left(\X_{n},Y_{n},Z_{n}\right)\right\},
\end{align*}
where each
 $(\X_k,Y_k)$ corresponds to a different observation taken from $\mathcal{D}^{(1)}\cup \mathcal{D}^{(2)}$  and $Z_k \in \{1,2\}$ indicates from which dataset $(\X_k,Y_k)$ comes from. We then randomly split $\mathcal{D}$ into two sets: $\mathcal{D}^{tr}$ (training set)  and $\mathcal{D}^{te}$ (test set).
We use $\mathcal{D}^{tr}$ to train a probabilistic classifier that predicts
$Z$. The features used to predict $Z$ are (i) $(\X,Y)$ to estimate the amount of total dataset shift ($\textup{KL}_{\X,Y}$), (ii) $\X$ to estimate the amount of feature shift ($\textup{KL}_{\X}$), and (iii) $Y$ to estimate the amount of response shift ($\textup{KL}_{Y}$).
The estimated Radon–Nikodym derivative in the case of total dataset shift (the other ones are analogous) is given by 
$$ \widehat{\frac{dP_{\X,Y}^{(2)}}{dP_{\X,Y}^{(1)}}}(\x,y):=\frac{n_{tr}^{(1)}}{n_{tr}^{(2)}}\frac{\widehat \P (Z=2|\X=\x,Y=y)}{\widehat\P (Z=1|\X=\x,Y=y)},$$
where $\widehat\P$ denotes the trained probabilistic classifier and $n_{tr}^{(i)}$ is the number of samples from population $i$ in $\mathcal{D}^{tr}$. If target-labeled samples are scarce, training the classifier $\widehat\P$ from scratch can be challenging when $\X$ is high-dimensional. However, if target unlabeled samples are abundant, one possible solution is first training a classifier only depending on $\X$ and then using $h(\x)=\widehat\P(Z=2|\X=\x)$ to reduce the dimensions of $\X$ before training the classifier dependent on both $\X$ and $Y$. This solution does not affect the reliability of our method since false alarm control is not affected.

Finally, we use empirical averages\footnote{The same approach is used to estimate divergences by \citet{sonderby2016amortised} in the context of generative models, for example.} over the test dataset $\mathcal{D}^{te}$
to estimate the KL divergence (again in the case of total dataset shift; the other ones are analogous): $$  \widehat{\textup{KL}}_{\X,Y} := \frac{1}{|\mathcal{D}^{te}_2|}  \sum_{(\X_k,Y_k,Z_k) \in \mathcal{D}^{te}_2}  \log \left( \widehat{\frac{dP_{\X,Y}^{(2)}}{dP_{\X,Y}^{(1)}}}(\X_k,Y_k)\right),$$ 
where $\mathcal{D}^{te}_2:=\{(\X_k,Y_k,Z_k) \in \mathcal{D}^{te}:Z_k=2\}$ denotes the test samples from the second population.

This approach however cannot be used to estimate $\textup{KL}_{\X|Y}$ or $\textup{KL}_{Y|\X}$. Instead, we rely on the KL divergence decomposition, given in the following proposition extracted from \citet{polyanskiy2022information} (Theorem 2.13). To keep this text as self-contained as possible, we include a proof in the appendix.

\begin{prop}\label{prop:addit}
   Let $\textup{KL}_{Y}$, $\textup{KL}_{X}$, $\textup{KL}_{Y|X}$, $\textup{KL}_{X|Y}$, $\textup{KL}_{X,Y}$ be defined as they were in Section \ref{sec:test_stat}. Then
\begin{align*}
\textup{KL}_{\X,Y}&=\textup{KL}_{Y|\X}+\textup{KL}_{\X}\\
&=\textup{KL}_{\X|Y}+\textup{KL}_{Y}
\end{align*} 
\end{prop}
This result shows that the KL divergences of the conditional distributions can be estimated via $\widehat{\textup{KL}}_{\X|Y}:=\widehat{\textup{KL}}_{\X,Y}-\widehat{\textup{KL}}_{Y}~~\text{ and }~~\widehat{\textup{KL}}_{Y|\X}:=\widehat{\textup{KL}}_{\X,Y}-\widehat{\textup{KL}}_{\X}.$

\subsection{Hypothesis Testing}\label{sec:hypo_test}

Once we have statistics that can quantify the magnitude of different types of dataset shifts, we can use them to formally test the hypotheses described in Section \ref{sec:test_stat}. In this section, $Y$ can be discrete or continuous,  except when obtaining the $p$-values for the hypothesis $P^{(1)}_{\X|Y}= P^{(2)}_{\X|Y}$, in which we assume it is discrete. This is needed since Algorithm 1, in the appendix, relies on this assumption. If $Y$ is continuous or has few repeated values, the conditional shift of type 1 can be tested by discretizing/binning the label for computing the statistic and applying the algorithm\footnote{Binning is not needed when training the classifiers though.} -- we give more details and references at the end of this section. 

Consider the datasets  $\mathcal{D}^{te}$ and $\mathcal{D}^{te}_2$ as defined in Section \ref{sec:test_stat}, and let $T(\mathcal{D}^{te}_2)$ be a test statistic of interest computed using $\mathcal{D}^{te}_2$. Namely, $T(\mathcal{D}^{te}_2)$ can represent any of the following random quantities, depending on which type of shift we are testing for: $\widehat{\textup{KL}}_{Y}$, $\widehat{\textup{KL}}_{X}$, $\widehat{\textup{KL}}_{Y|X}$, $\widehat{\textup{KL}}_{X|Y}$, $\widehat{\textup{KL}}_{X,Y}$. We test each of the hypotheses of interest by computing a $p$-value of the form
\begin{align}
\label{eq:pval}
p(\mathcal{D}^{te})=\frac{1+\sum_{j=1}^B \I\left[ T(\mathcal{D}^{te}_2) \leq T\left(\widetilde{ \mathcal{D}}^{{te}^{(j)}}_2\right)\right]}{B+1}, 
\end{align}
where each $\widetilde{ \mathcal{D}}^{{te}^{(j)}}_2$ is a modified version of 
$\mathcal{D}^{te}_2$, which can depend on the whole test set $\mathcal{D}^{te}$. The modification that is done depends on the hypothesis we are testing:

\begin{itemize}
    \item To test the hypotheses related to unconditional distributions ($H_{0,\text{D}},H_{0,\text{F}}$ and $H_{0,\text{R}}$), $\widetilde{ \mathcal{D}}^{{te}^{(j)}}_2$ is obtained randomly permuting $Z_k$'s on $\mathcal{D}^{te}$ and then selecting the samples with $Z_k=2$ to form the modified version of $\mathcal{D}^{te}_2$.  In this case, $p$ is the $p$-value associated with a permutation test, a method commonly used to perform two-sample tests \cite{lehmann2005testing}.
    \item To test $H_{0,\text{C1}}$, $\widetilde{ \mathcal{D}}^{{te}^{(j)}}_2$ is obtained randomly permuting the values of $Z_k$'s \emph{within each level of $Y$} on $\mathcal{D}^{te}$ and then selecting the samples with $Z_k=2$ to form the modified version of $\mathcal{D}^{te}_2$. Thus, we require $Y$ to be discrete to apply this test. In this case, $p$ is the $p$-value associated with a \emph{conditional independence} local permutation test \cite{kim2021local}.
    \item To test $H_{0,\text{C2}}$, we first estimate the conditional distribution of $Y|\X$ using the whole training set $\mathcal{D}^{tr}$ of labeled samples. Let  $Q_{Y | \X}$ denote such an estimate, which can be obtained using any probabilistic classifier, such as logistic regression, neural networks, CatBoost classifier \cite{prokhorenkova2017catboost}, or conditional density estimators \cite{izbicki2017converting} and GANs \cite{bellot2019conditional} if $Y$ is continuous. We then obtain $\widetilde{ \mathcal{D}}^{{te}^{(j)}}_2$ by replacing each $Y_k$ in $\mathcal{D}^{te}_2$ by a random draw from $Q_{Y | \X=\X_k}$. This test is known as the conditional randomization test (CRT) \cite{candes2018panning, berrett2020conditional}.
\end{itemize}

\begin{figure}[t]%
    \centering
    \includegraphics[width=.9\textwidth]{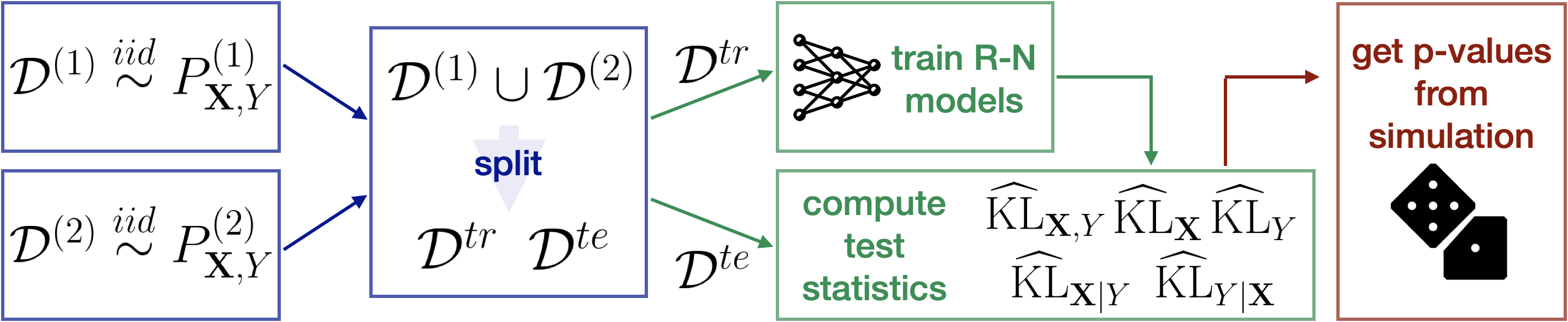} 
    \caption{\textit{DetectShift} in a nutshell. First, data are sampled from source and target distributions, mixed together, and split into a training and a test set. Then, using the training set, the Radon–Nikodym (R-N) models are trained, and all the test statistics are computed using the test set. Finally, $p$-values to conduct the tests are obtained through simulation procedures.}
    \label{fig:framework}
\end{figure}

Algorithm 1, in the appendix, details the steps to obtain the $p$-values for each case. For all the cases, we assume the procedure involving the training of probabilistic classifiers, described in Section \ref{sec:test_stat}, has already been executed. That is, we have a test statistic $T$ for every test we want to perform. Also, for the case we are testing for conditional shift type 2, we assume the estimated conditional distribution $Q_{Y | \X}$ has been computed. We fix a significance level $\alpha \in (0,1)$, and after calculating the $p$-value $p$ for a specific null hypothesis of interest, reject that hypothesis if $p\leq \alpha$. Figure \ref{fig:framework} summarizes how our framework, \textit{DetectShift}, works.

The following proposition shows that such tests are valid (that is, they control Type I error rate). In other words, false alarms are controlled. The only exception is the test for $H_{0,\text{C2}}$, which approximately controls Type I error probabilities as long as $Q_{Y | \X}$ is a good approximation of $P^{(2)}_{Y | \X}$. For the following result, we consider that $T$ and $Q_{Y | \X}$ are given and fixed, as both are obtained from the training set. Also, we make sure that $(T(\mathcal{D}^{te}_2), T(\widetilde{ \mathcal{D}}^{{te}^{(1)}}_2), \cdots, T(\widetilde{ \mathcal{D}}^{{te}^{(B)}}_2))$ has no repeated values by adding small centered Gaussian noises\footnote{This step is performed to break ties among data modifications, making it easy to show Type-I error is controlled.} to $T(\mathcal{D}^{te}_2)$ and $T(\widetilde{ \mathcal{D}}^{{te}^{(j)}}_2)$, for every $j$.
\begin{prop}\label{prop:false_alarms}
Let $p(\mathcal{D}^{te})$ be a $p$-value obtained from Algorithm 1 (appendix) with fixed $T$ and $Q_{Y | \X}$. Then, for every $\alpha \in (0,1)$,
\begin{itemize}
    \item Under
    $H_{0,\text{D}},H_{0,\text{F}}$,$H_{0,\text{R}}$, and $H_{0,\text{C1}}$ (if $Y$ is discrete),
    $$\P(p(\mathcal{D}^{te})\leq \alpha) \leq \alpha$$

    \item Under
    $H_{0,\text{C2}}$
        $$\P(p(\mathcal{D}^{te})\leq \alpha) \leq \alpha+\E_{\bar{P}^{(2)}_{\X}}\left[d_{\text{TV}}(\bar{Q}_{Y | \X},\bar{P}_{Y | \X})\right],$$
        where $ d_{\text{TV}}$ is the total variation distance, $\bar{P}^{(2)}_{\X}:=\prod_{(\X_k,Y_k,Z_k)\in \mathcal{D}_2^{te}}P^{(2)}_{\X_k}$, $\bar{Q}_{Y | \X}:=\prod_{(\X_k,Y_k,Z_k)\in \mathcal{D}_2^{te}} Q_{Y | \X=\X_k}$, and $\bar{P}_{Y | \X}=\prod_{(\X_k,Y_k,Z_k)\in \mathcal{D}_2^{te}} P^{(2)}_{Y | \X=\X_k}$.
\end{itemize}
\end{prop}

Here, $\prod$ denotes products of probability distributions. {Proposition \ref{prop:false_alarms} borrows well-studied results from the Statistics literature. The results for $H_{0,\text{D}}$, $H_{0,\text{F}}$, and $H_{0,\text{R}}$ are directly obtained by the fact that we use a permutation test \cite{lehmann2005testing}, while the result for $H_{0,\text{C1}}$ is obtained because of properties of local permutation \cite{kim2021local} and the result for $H_{0,\text{C2}}$ is obtained by adapting the results of \citet{berrett2020conditional} (Theorem 4) to our context. Suppose $Y$ is continuous or has few repeated values. In that case, if we discretize/bin it to test the conditional shift of type 1, our approach leads to an approximate test for $H_{0,\text{C1}}$ in the sense it approximately controls the Type I error. See \citet{kim2021local} (Theorems 2 and 3) for more details. To make this work self-contained, we include a proof for Proposition \ref{prop:false_alarms} in the appendix.

%% file: exp.tex
\section{Experiments}\label{sec:exp}

This section presents numerical experiments with both artificial and real data. In all the experiments in which $Y$ is discrete, we use the plug-in estimator (appendix) to estimate $\textup{KL}_Y$.

\subsection{Artificial Data Experiments}

\subsubsection{Detecting different types of shifts in isolation} 

\begin{wrapfigure}{r}{0.4\textwidth}
    \centering
    \vspace{-13pt}
    \includegraphics[width=.4\textwidth]{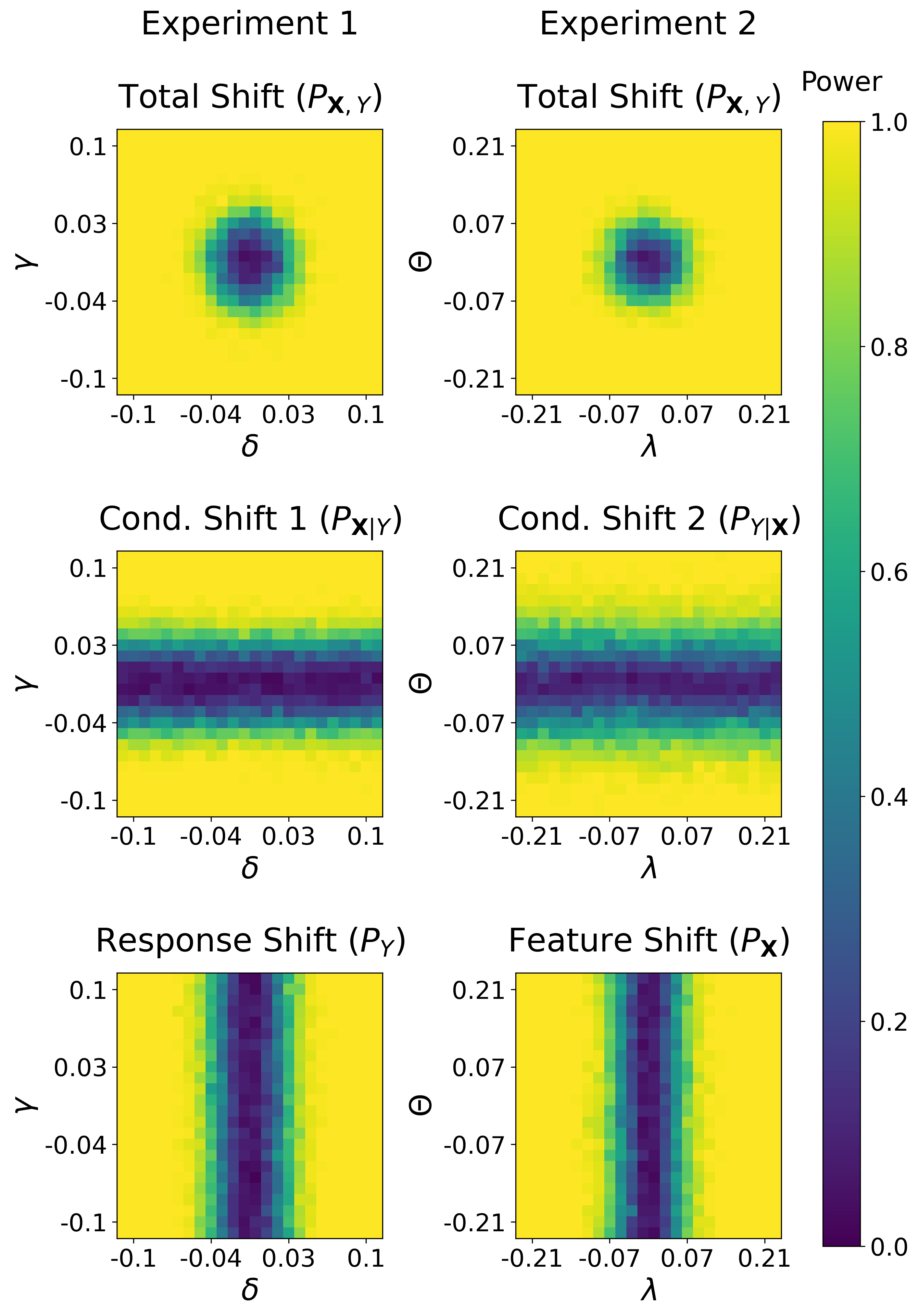}
    \caption{Power surfaces. In these experiments, $\delta$ controls response shift, $\gamma$ controls conditional shift 1, $\lambda$ controls feature shift, and $\theta$ controls conditional shift 2. While the tests can control Type I errors, their  power gets close to 1 when the shifts are bigger. Moreover, our procedure can detect the types of shifts in isolation. \textit{Therefore, the novel unified framework can reliably detect different types of dataset shifts.}} 
    \label{fig:power}
\end{wrapfigure}

In the first experiment, we set $$P^{(1)}_{Y}=\textup{Ber}(1/2),~~P^{(2)}_{Y}=\textup{Ber}(1/2+\delta)$$ and $$P^{(1)}_{X|Y}=\mathcal{N}(Y\cdot1_d,\textup{I}_d),~~P^{(2)}_{X|Y}=\mathcal{N}\left((Y+\gamma)\cdot1_d,\textup{I}_d\right),$$
where $\textup{Ber}(p)$ denotes the Bernoulli distribution with mean $p$, $1_d$ denotes a vector of ones of size $d=3$, $\textup{I}_d$ the identity matrix of dimension $d=3$, and $\mathcal{N}(\mu,\Sigma)$ denotes the normal distribution with mean vector $\mu$ and covariance matrix $\Sigma$. This way, $\delta$ controls the amount of response shift, while $\gamma$ controls the amount of conditional shift of type 1. Indeed, it is possible to show that $\textup{KL}_{Y}$ positively depends only on $|\delta|$ while $\textup{KL}_{X|Y}$ positively depends only on $|\gamma|$. In the second experiment, we set $$P^{(1)}_{X}=\mathcal{N}(0,1),~~P^{(2)}_{X}=\mathcal{N}(\lambda,1)$$ and  $$P^{(1)}_{Y|X}=\mathcal{N}(X,1),~~P^{(2)}_{Y|X}=\mathcal{N}(X+\theta,1),$$

where $\mathcal{N}(\mu,\sigma^2)$ denotes the normal distribution with mean $\mu$ and variance $\sigma^2$. In this way, $\lambda$ controls the amount of feature shift, while $\theta$ controls the amount of conditional shift of type 2. Indeed, it is possible to show that $\textup{KL}_{X}$ positively depends only on $|\lambda|$ while $\textup{KL}_{Y|X}$ positively depends only on $|\theta|$.

We vary $(\delta,\gamma)$ and $(\lambda,\theta)$ in a grid of points for experiments 1 and 2, respectively. For each point in the grid, we perform 100 Monte Carlo simulations to estimate the tests' powers, that is, the probabilities of rejecting the null hypotheses. For each pair $(\delta,\gamma)$ or $(\lambda,\theta)$ and Monte Carlo simulation, we: (i) draw training and test sets, from both joint distributions, with size 2500 each; (ii) train a logistic regression\footnote{We use Scikit-Learn's \cite{pedregosa2011scikit} default configuration with no hyperparameter tuning.\label{foot:scikit}} model as a probabilistic classifier to estimate the Radon-Nikodym derivatives using the training sets; (iii) use the test set from the target population to estimate $\textup{KL}_{X,Y},$ $\textup{KL}_{X|Y}$ or $\textup{KL}_{Y|X}$, and $\textup{KL}_{Y}$ or $\textup{KL}_{\X}$; (iv) and use the test set to calculate the $p$-values using\footnote{For the conditional randomization test, we train a linear regression model with Gaussian errors to estimate the conditional distribution of $Y$ given $X$ using the full training set.} Algorithm 1 in the appendix setting $B=100$; (v) reject the null hypothesis if the $p$-value is smaller than the level of significance $\alpha=5\%$.

Figure \ref{fig:power} shows the power estimates for each test as a function of $(\delta,\gamma)$ or $(\lambda,\theta)$. Our procedure to test the presence of different types of dataset shift is well-behaved: the power is close to the nominal level $\alpha=5\%$ when $(\delta,\gamma)$ or $(\lambda,\theta)$ is close to the origin, i.e., when no shift happens and grows to 1 when $||(\delta,\gamma)||$  or $||(\lambda,\theta)||$ gets larger.  Moreover, our procedure could also detect types of shifts in isolation: the power of our tests increases for conditional shift (types 1 and 2) and response/feature shift detection when increasing $|\gamma|$ or $|\lambda|$ and $|\delta|$ or $|\gamma|$ separately. As expected, the tests are not affected by the shifts that are not being tested at that  moment. 

In these experiments, we showed that the novel unified framework could reliably detect different dataset shifts in isolation.

\subsubsection{Comparisons with existing approaches} 

Now, we compare our framework with existing methods for detecting  shifts. We do this by comparing the power of the different hypotheses tests fixing $\alpha=5\%$. For this experiment, we use the same data-generating process used in the first two experiments and set the sample sizes of training and test sets to 500. Moreover, we use 200 Monte Carlo simulations to estimate power and set $B=100$ for Algorithm 1.  When our objective is to detect response and feature shifts, we vary $\delta$ and $\lambda$ but fix $\gamma=\theta=0$; when we aim to see both types of conditional shifts, we vary $\gamma$ and $\theta$ but set $\delta=\lambda=0$.

The main alternative approach we compare our method with is the total variation (TV) approach proposed by \citet{webb2018analyzing}, quantifying different types of dataset shift. To apply their method, we discretized the data and used our Algorithm 1 (appendix) to obtain $p$-values. When testing for response shift, we also include comparisons
with a Z-test to compare two proportions \cite{lehmann2005testing}, a $\chi^2$ test \cite{rabanser2019failing}, and a classification-based two-sample test \cite{lopez2016revisiting}. When testing for feature shift, we also include comparisons
with a Kolmogorov-Smirnov (KS) test \cite{kolmogorov1933, Smirnov39}, an MMD-based test \cite[Corollary 16]{gretton2012kernel}, and a classification-based two-sample test \cite{lopez2016revisiting}. For the classification-based two-sample tests, we use \citet{lopez2016revisiting}'s formulation to obtain the $p$-values. Finally, when we test for conditional shifts 1 and 2, we include two instances of the local permutation test\footnote{Permuting data within each level of $Y$.} (LPT) \cite{kim2021local} and the conditional randomization test (CRT) \cite{berrett2020conditional,candes2018panning,bellot2019conditional} using statistics based on the classification approach \cite{lopez2016revisiting}. More details can be found in the appendix. 
\begin{figure}[t]%
    \centering
    \includegraphics[width=.9\textwidth]{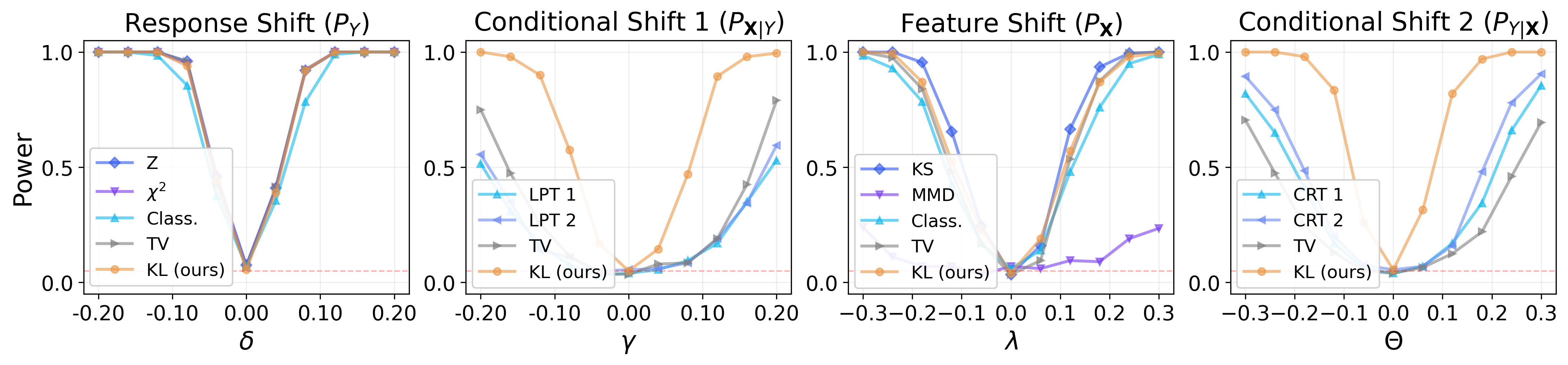}
    \caption{Comparing our framework with existing approaches plotting their power curves when $\alpha=5\%$. Our method had power curves similar to the alternative approaches when testing for response and feature shifts. However, our method achieved significantly higher power when testing both conditional shifts. \textit{In summary, (i) our approach  has the advantage of being unified, i.e., a single framework is used to test all hypotheses using test statistics with the same nature, and (ii) when compared to specialized tests, the novel unified framework does not lose in power when detecting marginal shifts while it wins when detecting conditional shifts}.
    }
    \label{fig:power2}
\end{figure}

Figure \ref{fig:power2} shows that our method had similar power curves to the alternative approaches when testing for response and feature shift. However, when testing for both types of conditional shift, our method achieved a significantly higher power when compared with the alternative approaches.

Next, we investigate the role of the dimensionality of the feature space in the performance of the three methods used to test for feature shift, and that can be easily extended to multidimensional cases. More specifically, our goal in the example is to detect feature shifts using the settings from the second experiment of this section when $\lambda=.24$. We concatenate to $X$ a standard Gaussian random vector (independent of the original $X$), ending up with an updated version of $X$ with size $d$. Then, we compare the various tests in terms of their power to test $H_{0,F}$ when $\alpha=5\%$. We use 200 Monte Carlo simulations to estimate power and set $B=100$ for Algorithm 1 (appendix). Because the divergence between the distributions remains the same when adding this noise, this experiment allows us to isolate the dimensionality.

\begin{wrapfigure}{r}{0.5\textwidth}
    \centering
    \includegraphics[width=.4\textwidth]{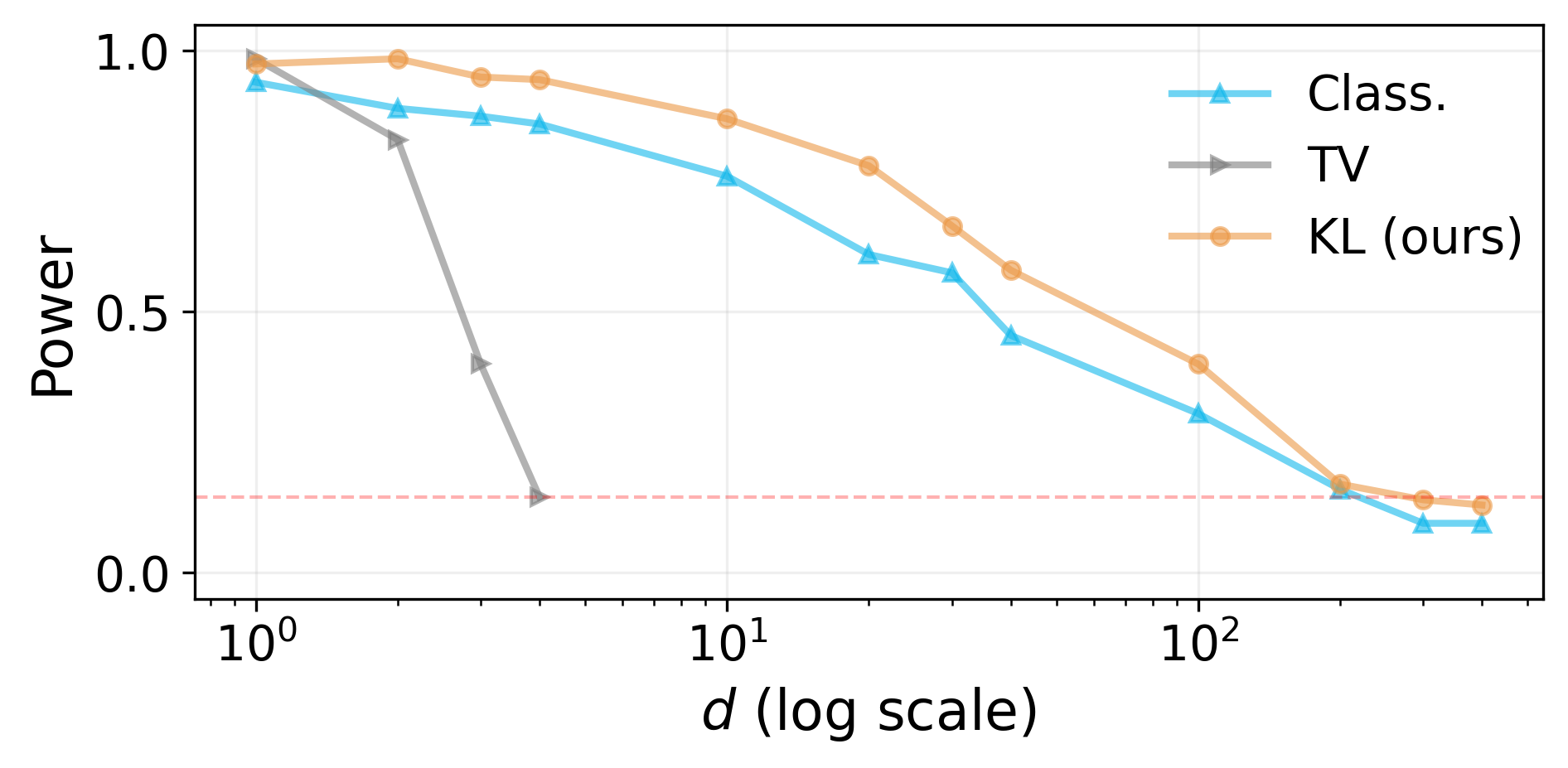}
    \caption{Role of the feature space's dimensionality in the tests' power.  The performance of our method  and the classification approach when $d=400$ is equivalent to the performance of the TV approach when $d=4$. \textit{Our method scales well to high-dimensional data and consistently outperforms  other approaches}.} 
    \label{fig:dimensions}
\end{wrapfigure}

Figure \ref{fig:dimensions}  indicates that the performance of our method and the classification approach does not suffer as much from increasing $d$ compared to the TV approach. We compare the TV approach with $d \in \{1,2,3,4\}$ with our approach $d \in \{1,2,3,4,10, \linebreak 20,30,40,100,200,300,400\}$. We stop at $d=4$ for the TV approach because the quantity of bins increases geometrically with the number of dimensions, and if $d\geq5$, we would expect to find less than two data points per bin. Interestingly, the performance of the TV approach when $d=4$ is equivalent to the performance of the other methods when $d=400$. Moreover, our method consistently outperforms the other two approaches.

In summary, (i) our approach has the advantage of being unified, i.e., a single framework is used to test all hypotheses using test statistics with the same nature while maintaining good power; (ii) our method scales well to high-dimensional data and consistently outperforms other approaches.

\subsection{Real Data Experiments}

\subsubsection{Insights from credit data}

In this experiment, we use our method to extract insights into how probability distributions can differ in a financial application. The dataset used in this experiment is a credit scoring dataset and was kindly provided by the Latin American Experian DataLab, based in Brazil. It contains financial data of one million Brazilians collected every month going from August/2019 to May/2020.

The features in this dataset are related to past financial data, e.g., amount of loans and credit card bills not paid on time, and the \emph{label variable} informs whether a consumer will \emph{delay a debt payment} for 30 days in the \emph{next three months}, i.e., we have a binary classification problem. In this experiment, we kept 20k random data points each month, with $20\%$ of them going to the test set. Also, we kept the top 5 most essential features of the credit risk prediction model. These specific features are related to payment punctuality for credit card bills, the number of active consumer contracts, and the monetary values involved.  We used CatBoost \cite{prokhorenkova2017catboost} both to estimate the Radon-Nykodim derivative and the conditional distribution of $Y|\X$.

\begin{wrapfigure}{r}{0.5\textwidth}
    \centering
    \includegraphics[width=.4\textwidth]{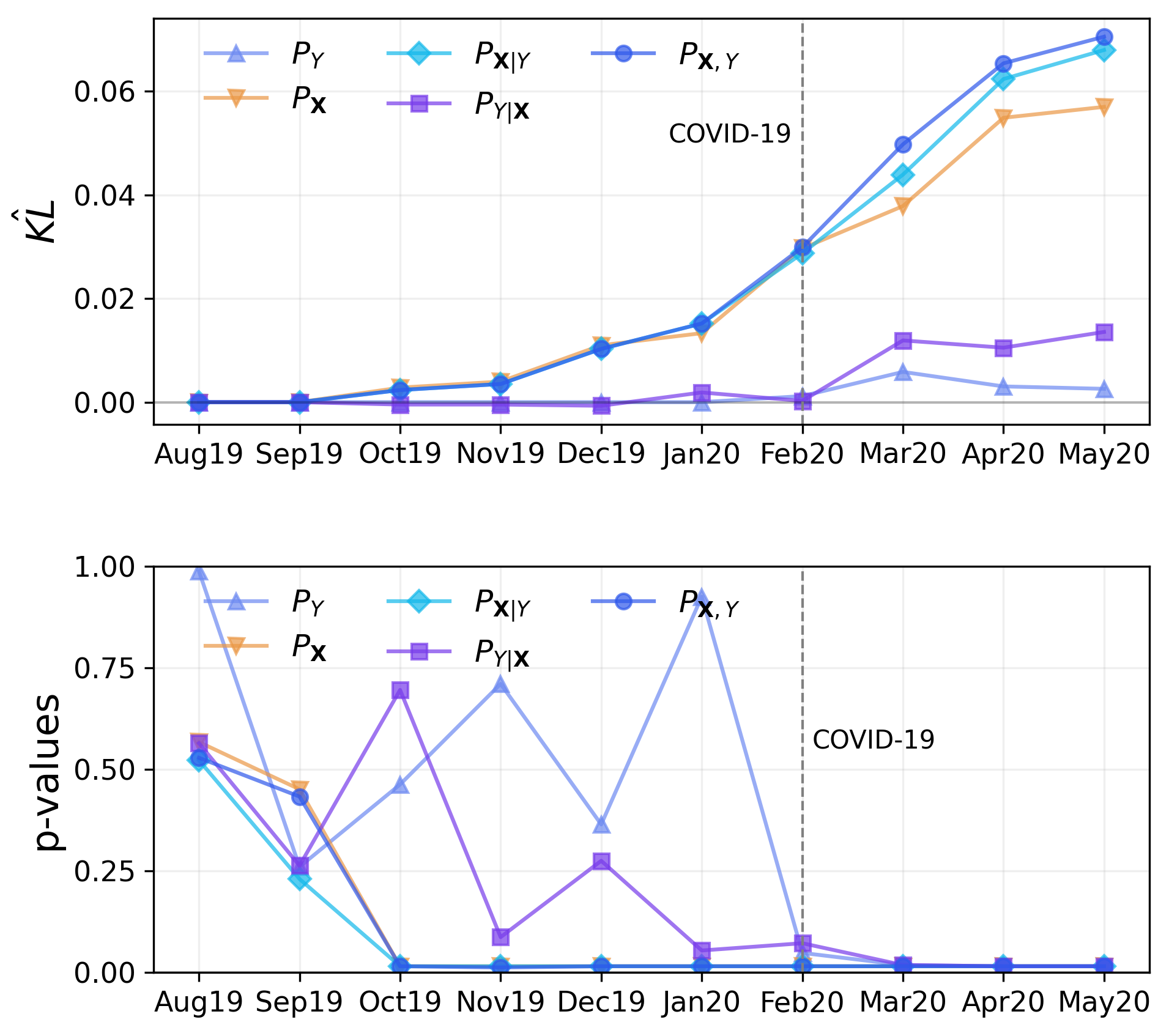}
    \caption{Detecting different types of dataset shift using credit data from the Latin American Experian DataLab, based in Brazil. The vertical dashed line marks the beginning of the COVID-19 crisis in Brazil. We highlight the decoupling between the total shift and covariate or conditional shift type 1 after February/2020. This behavior is due to a bigger shift in the marginal and conditional distribution of $Y$ and is possibly associated with the economic consequences of the pandemic. \textit{Because all tests use test statistics with the same nature, we can easily compare the magnitude of the different types of shifts. The testing procedure becomes interpretable.}} 
    \label{fig:credit_exp}
\end{wrapfigure}

The results in Figure \ref{fig:credit_exp} indicate increasing covariate, conditional shift type 1, and total dataset shift from the beginning. This is expected because the features contain information about how people use their credit (e.g., loan amount, credit card use). The way people use their credit is a function of changes in the economy that can occur rapidly, such as fluctuating inflation/interest/exchange rates, extra expenses due to holidays, etc. From February 2020 onward, i.e., labels relative to months post-March 2020, it is possible to notice a decoupling between shift curves in the coming months after the first official COVID-19 case detected in Brazil and the beginning of the economic crisis. The decoupling means that a more significant share of the total shift is due to a shift in the marginal and conditional distribution of $Y$. We speculate that this decoupling is due to measures taken by banks and credit bureaus to help consumers during the pandemic. Some measures include but are not limited to, longer payment intervals and lower interest rates. 

To conclude, because all test statistics have the same nature, we can easily compare the magnitude of the different types of shifts. The testing procedure becomes interpretable.

\subsubsection{Using dataset shift diagnostics to improve predictions} In this experiment, we evaluate our method as a guide for dataset shift adaptation using the MNIST and USPS datasets \cite{lecun1998gradient, xu2021concept}. Both datasets contain images, i.e., pixel intensities, and labels for the same ten digits (i.e. 0 to 9).  Our interest is (i) to use our framework to quantify and formally test the presence of all types of dataset shift using the MNIST distribution as the source and a mixture between MNIST and USPS (with increasing proportions of USPS participation) as target distributions and then (ii) adapt our predictors using the insights given by our diagnostics to achieve better out-of-sample performance. We aim to show how detecting specific types of shift help practitioners correct their models. In this experiment, we use $256$ features (pixel intensities of $16\times 16$ images) and partition the data in 12 smaller disjoint datasets of size 3.1k in a way that the proportion of USPS samples increases linearly from $0\%$ to $50\%$.  We split each dataset with $10\%$ of the samples to test and use CatBoost \cite{prokhorenkova2017catboost} to estimate R-N derivatives and conditional distribution. We use the first dataset, with only MNIST samples, as our baseline and compare it to the mixed datasets. 

\begin{figure}[t]
    \centering
    \hspace*{-.4cm}
    \includegraphics[width=.85\textwidth]{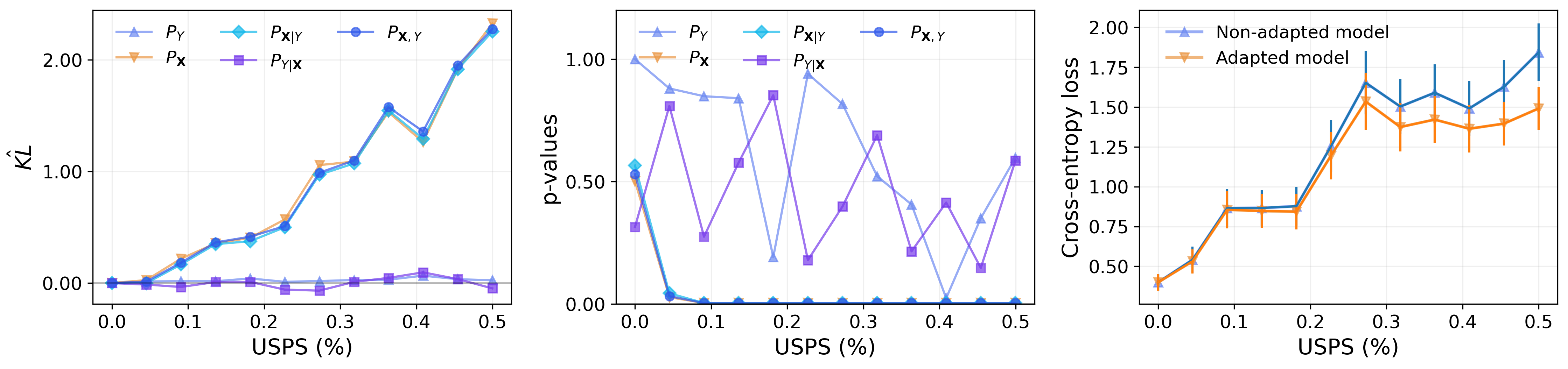}
    
    \caption{Detecting different types of dataset shift using MNIST and USPS data and then adapting a predictor. We define the MNIST distribution (USPS $(\%)=0$) as the source distribution and the mixed distributions (USPS $(\%)>0$) as multiple target distributions. Response and conditional shifts of type 2 (shift in $P_{Y|\X}$) are not evident, while others are. The second plot indicates that adapting for feature shift should be enough (as $P_{Y|\X}$ seems static). In the third plot, we compare the performance of two logistic regression models trained using pure MNIST samples with one adapted for feature shift. The adapted model gives better predictions. \textit{Thus, by properly leveraging the insights given by our framework, one can improve the predictive power of the supervised learning model.}}
    
    \label{fig:digits_exp}
\end{figure}

The plots in Figure \ref{fig:digits_exp} indicate that the total dataset shift (shift in $P_{\X,Y}$), conditional shift 1 (shift in $P_{\X|Y}$), and feature shift (shift in $P_{\X}$) are promptly detected while response shift (shift in $P_{Y}$) and conditional shift 2 (shift in $P_{Y|\X}$) are not. That observation is consistent with the fact that (i) the  distribution of $Y$ is similar in both MNIST and USPS populations which also implies that (ii) two similar pixel configurations should induce similar posterior distributions of labels regardless of the origin distribution. Finally, the plots indicate that adapting for feature shifts should be enough to achieve better predictions on the target domain. Indeed, that is the case here -- in the third plot, we compare the performance of two logistic regression models\footref{foot:scikit}. Both are trained using pure MNIST samples, but one is correct for feature shift using importance weighting \cite{sugiyama2007covariate}. \textit{The weights are obtained via the classifier used to estimate $\text{KL}_\X$, with no need to fit an extra model for the weights.}

In this experiment, we show that, by properly leveraging the insights given by our framework, one can improve the predictive power of the supervised learning model.

\subsubsection{Detecting shifts with deep models} We use our framework to detect shifts in image and text datasets using deep learning models as classifiers to estimate the Radon-Nikodym derivative and the conditional distribution of $Y|\X$. We use large pre-trained models as feature extractors, freezing all the layers except the output one, given by a logistic regression model. The pre-trained models are VGG-16 \cite{simonyan2014very} for images and XLM-ROBERTa \cite{conneau2020unsupervised} for texts. We use the Tiny ImageNet \cite{deng2009imagenet,abai2019densenet} and CIFAR-10 \cite{krizhevsky2009learning} as image\footnote{Tiny ImageNet contains $64\times 64$ images from 200 classes while CIFAR-10 contains $32\times 32$ images from 10 classes.} datasets  and “Amazon Fine Food Reviews,” available on Kaggle, as our text dataset. Then, the first two datasets we use are composed of RGB images from $K=10$ different classes\footnote{We group, in increasing order, the classes from Tiny ImageNet in 10 meta-classes.}. In contrast, the third dataset is composed of product reviews in the form of short texts and a rating, varying from 0 to 4, given by consumers, thus having $K=5$ classes\footnote{Regarding the Amazon dataset, we subsampled the data to guarantee all the classes have roughly the same number of examples.}. The sample sizes are 30k data points with $10\%$  going to test.

We derive the source and target datasets from the original datasets as follows. First, we fix $\delta \in (0,.5)$ and then create a list \code{LIST} of $K$ numbers (one for each class) where the first element of the list is $\delta$, the last is $1-\delta$, and the intermediate ones are given by linear interpolation of $\delta$ and $1-\delta$. Then, for each $k \in \{0,...,K-1\}$, we randomly select \code{LIST[}$k$\code{]} of the samples of class $k$ to be in the source dataset, while the rest goes to the target dataset. In this way, we explicitly introduce label shift (shift in $P_Y$ but not in $P_{\mathbf{X}|Y}$), which is expected to induce feature shift and conditional shift type 2 as well. After we have data from both populations (source and target), we detect the shifts as usual. We repeated the same procedure for all $\delta\in\{.5,.45,.4,.35,.3\}$ and for five different random seeds. 

\begin{figure*}[t]
    \centering
    \includegraphics[width=1\textwidth]{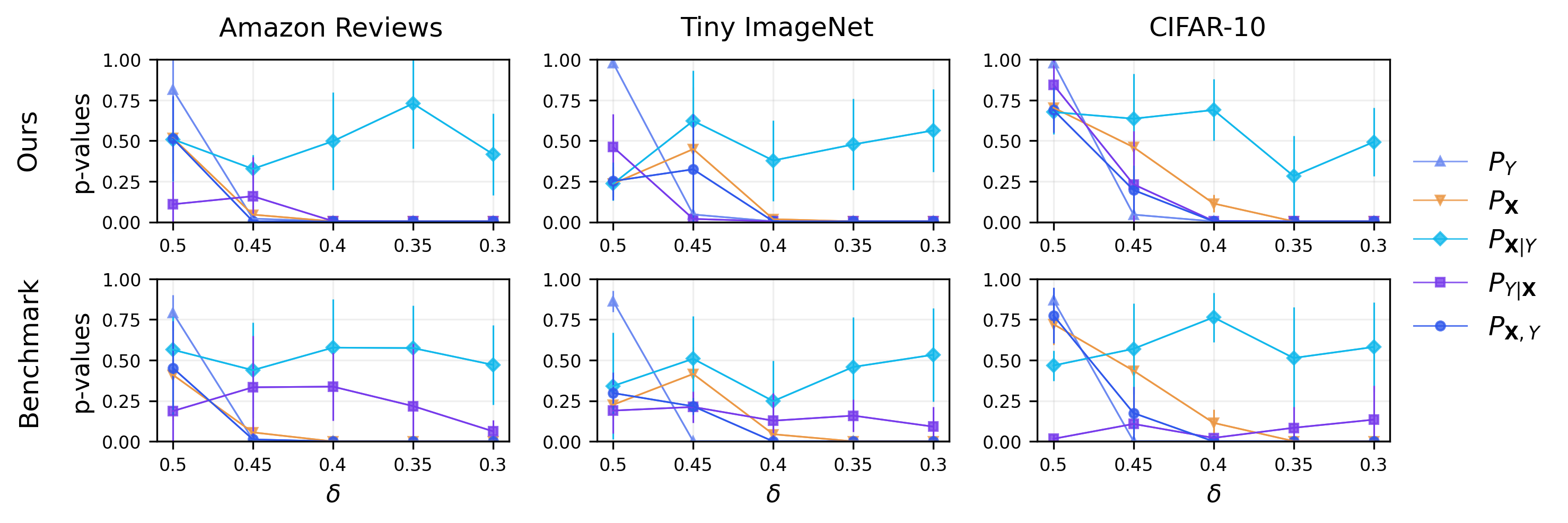}
    \vspace{-10pt}
    \caption{Detecting different types of dataset shift using complex high-dimensional data. For many values of $\delta$, we could detect all types of shift except conditional shift 1, which is expected because we introduced response shift in isolation ($\delta\neq.5$ is equivalent to label shift). The benchmark approaches cannot detect a conditional shift of type 2, which can be explained by lower power. \textit{Then, our framework can accurately detect the correct shifts even in complex datasets  and using deep models}.}
    
    \label{fig:deep}
\end{figure*}

The results are shown in Figure \ref{fig:deep}, where the lines represent averages across repetitions and error bars give the standard deviations. We were able to detect\footnote{Except when $\delta$ is close to $.5$ (small or no shift).} all types of shift except conditional shift 1 using our approach. This result was expected because, given the class, the distribution of the features must not be affected by how we introduced the shift. We also run results considering the classification-based two-sample test proposed by \cite{lopez2016revisiting} to test shifts related to joint/marginal distributions and CRT 1 and LPT 1, explained in detail in the appendix, to test for conditional shifts. For the benchmarks, we cannot detect a conditional shift of type 2. A lower power of the benchmarks compared to our method can explain that fact (see artificial data experiments). Furthermore, we cannot compare shift magnitudes using benchmark approaches since their statistics do not have the same nature. We report KL estimates of \textit{DetectShift} in the appendix. We  also left in the appendix an extra (and similar) experiment in which our framework obtained similar p-values compared to the benchmarks.

In summary, we show that our framework can accurately detect the correct shifts even in complex datasets and using deep models.

%% file: conclusions.tex
\section{Discussion}

The proposed \textit{DetectShift} framework quantifies and formally tests different dataset shifts, including response, feature, and conditional shifts.  Our approach sheds light not only on \emph{if} a prediction model should be retrained/adapted but also on \emph{how}, enabling the practitioners to tackle shifts in their data objectively and efficiently. Our method is versatile and applicable across various data types and supervised tasks. Unlike our framework, existing dataset shift detection methods are only designed to detect specific types of shift or cannot formally test their presence, sometimes even requiring both labels and features to be discrete or low-dimensional. 

Our experiments have provided compelling evidence of the effectiveness of our framework. Through artificial data experiments, we have demonstrated the versatility of \textit{DetectShift} in isolating different types of dataset shifts and leveraging supervised learning to construct powerful tests. Furthermore, our framework has shown remarkable performance in higher dimensions, surpassing possible benchmarks. In real data experiments, we have illustrated how \textit{DetectShift} can extract valuable insights from data, assisting practitioners in adapting their models to changing distributions. Moreover, our approach seamlessly integrates with deep learning models, making it applicable to real-world scenarios. These findings highlight the significant contributions of our framework, showcasing its efficacy in detecting dataset shifts and underscoring its potential for practical applications in various domains.

\section{Final remarks and possible extensions}\label{sec:final}

In conclusion, we reflect on key facets and potential limitations inherent to our framework.

\textbf{Why and how to obtain better Radon-Nikodym derivatives estimates?} While Type I error control is not affected by Radon-Nikodym derivatives estimation (as the theoretical results show), the power of the tests is. This happens because the classifiers' performance in predicting probabilities directly influences how well we approximate the KL divergence and, consequently, can detect shifts in the data. Thus, we can use the cross-entropy (CE) loss on a validation set to pick the best classifier.

\textbf{Modular framework and alternative statistics choices.} Our hypothesis tests are agnostic to the choice test statistics; therefore, our framework is modular. We consider the KL statistics desirable, among other things, because we have the additivity property (Proposition \ref{prop:addit}), making the analyses more interpretable. A natural alternative to the KL statistics, when we do not want to favor one of the distributions, is the symmetrized KL, for example.

\textbf{Dealing with streaming data.} While the primary design of our framework does not cater to streaming data, it possesses the flexibility for adaptation in such settings. One way of doing this is to group the data into batches (e.g., every hour, day, month) and then apply the proposed approach to compare two or more data batches. Multiple testing methods \cite{rice2008methods} can be used along with our framework if the practitioner desires.

\textbf{Labeled data in the target domain.} Even though there are many situations in which at least some labeled data are available for the target domain, there are cases in which that is not true. We recognize this scenario as a limitation, warranting further exploration in subsequent research.

\section{Acknowledgments}

We sincerely thank the ``Information Sciences'' editor and anonymous reviewers for their invaluable contributions to this paper. We are grateful for their time and dedication in carefully evaluating our manuscript and providing valuable suggestions for improvement. 

We are thankful for the credit dataset provided by the Latin American Experian Datalab, the Serasa Experian DataLab. We want to thank the developers of Grammarly and ChatGPT, which were helpful tools when polishing our text. 

FMP is grateful for the financial support of CNPq (32857/2019-7) and Fundunesp/Advanced Institute
for Artificial Intelligence (AI2) (3061/2019-CCP) during his master's degree at the University of São Paulo (USP). Part of this work was written when FMP was at USP.

RI is grateful for the financial support of CNPq (309607/2020-5 and 422705/2021-7) and FAPESP (2019/11321-9).

\section{Code and data}
Our package for dataset shift detection can be found in \url{https://github.com/felipemaiapolo/detectshift}

The source code and data used in this paper can be found in \url{https://github.com/felipemaiapolo/dataset_shift_diagnostics}.

%% file: append.tex
\section{Methodology}

\subsection{Alternative estimator for $\textup{KL}_Y$ when $Y$ is discrete}
 Assume $\mathcal{Y}$ is the range of $Y$ in the target domain, where $\mathcal{Y}$ is finite. Define $p_{y}^{(i)}=P_{Y}^{(i)}(\{y\})$, for $i=1,2$. Then, we can write $\textup{KL}_Y=\sum_{y \in \mathcal{Y}}p_{y}^{(2)}\log\frac{p_{y}^{(2)}}{p_{y}^{(1)}} $. Having observed two datasets (in practice, we use the test datasets), $\mathcal{D}^{(1)}$
and $\mathcal{D}^{(2)}$, we define $\hat{p}_{y}^{(i)}$ to be the relative frequency of the label $y$ in dataset $i$. Then, a plug-in estimator for $\textup{KL}_Y$ is given by $\widehat{\textup{KL}}_Y=\sum_{y \in \mathcal{Y}}\hat{p}_{y}^{(2)}\log\frac{\hat{p}_{y}^{(2)}}{\hat{p}_{y}^{(1)}} $. This estimator is consistent.

\subsection{Algorithm to obtain $p$-values}

\begin{small}
\begin{algorithm}[H]
\DontPrintSemicolon
 \KwIn{(i) Hypothesis to be tested and respective test statistic $T$, (ii) Test set $\mathcal{D}^{te}$, (iii) number of iterations $B \in \mathbb{N}$, (iv) conditional distribution $Q_{Y|\X}$ (in case of testing for conditional shift type 2), (v) Gaussian noise variance $\nu$;}

 \KwOut{p-value $p=p(\mathcal{D}^{te})$;}

 \medskip
 Initialize $C=0$ and obtain $\mathcal{D}^{te}_2=\{(\X_k,Y_k,Z_k) \in \mathcal{D}^{te}:Z_k=2\}$;

 Get $T_0 = T(\mathcal{D}^{te}_2) + N(0,\nu)$;
 
 \For{$j$ in $\{1,...,B\}$}
 {
 \medskip
 \uIf{Testing for response shift, feature shift, or dataset shift}{
    Draw a random permutation $\pi=(\pi_1~...~\pi_{|\mathcal{D}^{te}|})$ of natural numbers from $1$ to $|\mathcal{D}^{te}|$\;
    Set $\widetilde{\mathcal{D}^{te}_2}^{(j)}=\{(\X_k,Y_k,Z_{\pi_k}): (\X_k,Y_k,Z_k) \in \mathcal{D}^{te} \text{ and } Z_{\pi_k}=2\}$ \;
  }\medskip
  \uElseIf{Testing for conditional shift (type 1)}{
    Let $\mathcal{Y}$ be a finite set which $Y$ takes values\;
    \For{$y$ in $\mathcal{Y}$}
    {
        Get $\mathcal{D}^{(y)}=\{(\X_k,Y_k,Z_k)\in \mathcal{D}^{te}: Y_k=y\}$ \;
        Draw a random permutation $\pi^{(y)}=(\pi^{(y)}_1~...~\pi^{(y)}_{|\mathcal{D}^{(y)}|})$ of natural numbers from $1$ to $|\mathcal{D}^{(y)}|$\;
    }
    Set $\widetilde{\mathcal{D}^{te}_2}^{(j)}=\bigcup_{y \in \mathcal{Y}}\{(\X_k,Y_k,Z_{\pi^{(y)}_k}): (\X_k,Y_k,Z_k) \in \mathcal{D}^{(y)} \text{ and } Z_{\pi^{(y)}_k}=2\}$
  }\medskip
  \uElseIf{Testing for conditional shift (type 2)}{
    Sample $\widetilde{Y}_k|\X_k\sim Q_{Y|\X=\X_k}$, for each $\X_k$ from elements of $\mathcal{D}_2^{te}$ \;
    Set $\widetilde{\mathcal{D}^{te}_2}^{(j)}=\{(\X_k,\widetilde{Y}_k,Z_k): (\X_k,Y_k,Z_k) \in \mathcal{D}_2^{te}\}$ \;
  }\medskip
  Get $T_j = T(\widetilde{ \mathcal{D}^{te}_2}^{(j)}) + N(0,\nu)$;

  Update $C=C+\I\left[ T_0 \leq T_j\right]$;
}

$p=\frac{C+1}{B+1}$

\textbf{return} $p$.
\caption{\textit{DetectShift}: obtaining p-values}
\label{alg:pval}
\end{algorithm}
\end{small}

In practice, $\nu>0$ can be a very small number, e.g., $10^{-10}$.

\subsection{Proofs}

The results presented in this section are not original; however, we decided to include proofs to make this work self-contained.

\textbf{Proposition \ref{prop:addit}.} Let $\textup{KL}_{Y}$, $\textup{KL}_{X}$, $\textup{KL}_{Y|X}$, $\textup{KL}_{X|Y}$, $\textup{KL}_{X,Y}$ be defined as they were in Section \ref{sec:test_stat}. Then
\begin{align*}
\textup{KL}_{\X,Y}&=\textup{KL}_{Y|\X}+\textup{KL}_{\X}\\
&=\textup{KL}_{\X|Y}+\textup{KL}_{Y}
\end{align*}
\begin{proof}

    In this proof, we assume the distributions are absolutely continuous with respect to the Lebesgue measure. For a more general proof, see \citet{polyanskiy2022information} (Theorem 2.13).
    
    We start showing $\textup{KL}_{\X,Y}=\textup{KL}_{Y|\X}+\textup{KL}_{\X}$. Let $p^{(1)}_{\X,Y}$ and $p^{(2)}_{\X,Y}$ be the joint densities of $P^{(1)}_{\X,Y}$ and $P^{(2)}_{\X,Y}$ with respect to the Lebesgue measure. Then
    \begin{align*}
    \textup{KL}_{\X,Y}&=\int p^{(2)}_{\X,Y}(\x,y) \log\frac{p^{(2)}_{\X,Y}(\x,y)}{p^{(1)}_{\X,Y}(\x,y)} d(\x,y)\\
    &= \int p^{(2)}_{Y|\X}(y| \x)p^{(2)}_{\X}(\x) \log\frac{p^{(2)}_{Y|\X}(y| \x)p^{(2)}_{\X}(\x) }{p^{(1)}_{Y|\X}(y| \x)p^{(1)}_{\X}(\x) } d(\x,y)\\
    &= \int p^{(2)}_{\X}(\x) \int p^{(2)}_{Y|\X}(y| \x) \log\frac{p^{(2)}_{Y|\X}(y| \x)}{p^{(1)}_{Y|\X}(y| \x)}dy d\x+\int p^{(2)}_{\X}(\x) \log\frac{p^{(2)}_{\X}(\x) }{p^{(1)}_{\X}(\x) } d\x\\
    &= \textup{KL}_{Y|\X}+\textup{KL}_{\X}
    \end{align*}
    We can show that $\textup{KL}_{\X,Y}=\textup{KL}_{\X|Y}+\textup{KL}_{Y}$ analogously.
\end{proof}

\textbf{Proposition \ref{prop:false_alarms}.} Let $p(\mathcal{D}^{te})$ be a $p$-value obtained from Algorithm 1 (appendix) with fixed $T$ and $Q_{Y | \X}$. Then, for every $\alpha \in (0,1)$,
\begin{itemize}
    \item Under
    $H_{0,\text{D}},H_{0,\text{F}}$,$H_{0,\text{R}}$, and $H_{0,\text{C1}}$ (if $Y$ is discrete),
    $$\P(p(\mathcal{D}^{te})\leq \alpha) \leq \alpha$$

    \item Under
    $H_{0,\text{C2}}$
        $$\P(p(\mathcal{D}^{te})\leq \alpha) \leq \alpha+\E_{\bar{P}^{(2)}_{\X}}\left[d_{\text{TV}}(\bar{Q}_{Y | \X},\bar{P}_{Y | \X})\right],$$
        where $ d_{\text{TV}}$ is the total variation distance, $\bar{P}^{(2)}_{\X}:=\prod_{(\X_k,Y_k,Z_k)\in \mathcal{D}_2^{te}}P^{(2)}_{\X_k}$, $\bar{Q}_{Y | \X}:=\prod_{(\X_k,Y_k,Z_k)\in \mathcal{D}_2^{te}} Q_{Y | \X=\X_k}$, and $\bar{P}_{Y | \X}=\prod_{(\X_k,Y_k,Z_k)\in \mathcal{D}_2^{te}} P^{(2)}_{Y | \X=\X_k}$.
\end{itemize}
\begin{proof}
    We start deriving the result for $H_{0,\text{F}}$ using the theory behind permutation tests for two-sample/independence tests (see \citet[Chapter 15]{lehmann2005testing} for more details). The results for $H_{0,\text{D}}$ and $H_{0,\text{R}}$ obtained analogously. Define $\widetilde{\mathcal{D}}^{{te}^{(0)}}_2:=\mathcal{D}^{{te}}_2$ and $T_j := T\left(\widetilde{ \mathcal{D}}^{{te}^{(j)}}_2\right) + N_j$, for all $j$, where $N_j$'s are i.i.d. $N(0,\nu)$. For the rest of the proof, $T_j$'s will always be noisy versions of the respective  $T\left(\widetilde{ \mathcal{D}}^{{te}^{(j)}}_2\right)$'s.
    
    Recall that
    \begin{align*}
    p(\mathcal{D}^{te})&=\frac{\sum_{j=0}^B \I\left[ T_0 \leq T_j\right]}{B+1}
    \end{align*}
    See that $(B+1)p(\mathcal{D}^{te})$ gives the rank of $T_0$ among all $T_j$'s, that is, if $(B+1)p(\mathcal{D}^{te})=k$ it means that there are $k$ values of $j$ such that $T_0 \leq T_j$. If $P^{(1)}_\X=P^{(2)}_\X$ and the permutations are drawn uniformly from the set of all permutations of $\{1,\cdots,|\mathcal{D}^{te}|\}$, then $T_0,\cdots,T_B$ are exchangeable. Because we add $N_j$'s in $T_j$'s, we can guarantee that all $T_j$'s are different with probability 1. Consequently, $U_{B+1}:=(B+1)p(\mathcal{D}^{te})$ is uniformly distributed in $\{1,\cdots,B+1\}$. Then,
    \begin{align*}
    \mathbb{P}\left(p(\mathcal{D}^{te})\leq\alpha\right)&=\mathbb{P}\left((B+1)p(\mathcal{D}^{te})\leq (B+1)\alpha\right)\\
    &=\mathbb{P}\left(U_{B+1}\leq (B+1)\alpha\right)\\
    &=\mathbb{P}\left(U_{B+1}\leq  \left \lfloor (B+1)\alpha\right \rfloor\right)\\
    &=\frac{\left \lfloor (B+1)\alpha\right \rfloor}{B+1}\\
    &\leq \frac{ (B+1)\alpha}{B+1}
    \\
    &= \alpha
    \end{align*}

    where $\left \lfloor \cdot \right \rfloor$ is the floor function.
    
    The result for $H_{0,\text{C1}}$ is obtained similarly. Assume that $P^{(1)}_{\X|Y=y}=P^{(2)}_{\X|Y=y}$ for every $y\in\textup{supp}(P^{(2)}_Y)$. We can ignore values of $y$ not in $\textup{supp}(P^{(2)}_Y)$ because $p(\mathcal{D}^{te})$ does not depend on them. Given that we permute samples within each level of $Y$, we have that $T_0,\cdots,T_B$ are exchangeable, implying that (by our derivations above)
    
    \begin{align*}
    \mathbb{P}\left(p(\mathcal{D}^{te})\leq\alpha\right)\leq \alpha.
    \end{align*}

    The result for $H_{0,\text{C2}}$ is based on Theorem 4 of \citet{berrett2020conditional}. For each $j \in \{1,\cdots, B\}$, denote $\widetilde{\mathcal{D}^{te}_2}^{(j)}=\{(\X_k,\widetilde{Y}_k,Z_k): (\X_k,Y_k,Z_k) \in \mathcal{D}_2^{te}\}$ where $\widetilde{Y}_k|\X_k\sim Q_{Y|\X=\X_k}$, for each $\X_k$ from elements of $\mathcal{D}_2^{te}$.  Define $\widetilde{\mathcal{D}^{te}_2}^{(0)}:=\mathcal{D}^{te}_2$ and let $\widetilde{\mathcal{D}^{te}_2}^{(B+1)}$ be one extra dataset built in the same way as all others  $\widetilde{\mathcal{D}^{te}_2}^{(j)}$ considering $j \in \{1,\cdots, B\}$. See that all datasets $\widetilde{\mathcal{D}^{te}_2}^{(j)}$ considering $j \in \{0,\cdots, B+1\}$ share the same values for the covariates; let $\mathbb{X}$ denote a random matrix containing such values for all samples. Because the p-values only depend on the data coming from\footnote{If that was not true, the following inequality would hold if we condition not only on $\mathbb{X}$ but also on a matrix composed of $Z$'s (as done in \citet{berrett2020conditional}).} $\mathcal{D}_2^{te}$ (and not on the full test set $\mathcal{D}^{te}$), we have that

    \begin{align*}
    \mathbb{P}\left(p(\mathcal{D}^{te})\leq\alpha | \mathbb{X}\right) \leq \mathbb{P}\left(\Tilde{p}(\mathcal{D}^{te})\leq\alpha  | \mathbb{X}\right) + d_{\text{TV}}(\bar{Q}_{Y | \X},\bar{P}_{Y | \X})
    \end{align*}

    where $\Tilde{p}(\mathcal{D}^{te})$ is the p-value calculated using $\widetilde{\mathcal{D}^{te}_2}^{(B+1)}$ instead of $\widetilde{\mathcal{D}^{te}_2}^{(0)}$. This step is justified by the definition of the total variation distance and by a conditional independence argument \citep{berrett2020conditional} (given $\mathbb{X}$, all data replications are conditionally independent).
    
    By construction, $T_1,\cdots,T_{B+1}$ are exchangeable given $\mathbb{X}$, we have that $ \mathbb{P}\left(\Tilde{p}(\mathcal{D}^{te})\leq\alpha  | \mathbb{X}\right)\leq \alpha$. Then, taking expectations on both sides of the inequality above, we get
    \begin{align*}
    \mathbb{P}\left(p(\mathcal{D}^{te})\leq\alpha \right) \leq \alpha + \E_{\bar{P}^{(2)}_{\X}}\left[d_{\text{TV}}(\bar{Q}_{Y | \X},\bar{P}_{Y | \X})\right]
    \end{align*}
\end{proof}

    Realize that the last result is true even if $H_{0,\text{C2}}$ does not hold. However, if $H_{0,\text{C2}}$ is not true, we do not expect $\E_{\bar{P}^{(2)}_{\X}}\left[d_{\text{TV}}(\bar{Q}_{Y | \X},\bar{P}_{Y | \X})\right]$ to be small.

\section{Experiments}

\subsection{Comparisons with existing approaches (more details)}

Some details about the experiments were omitted from the main text: (i) how we choose $Q_{Y|\X}$ when testing for conditional shift 2; (ii) how the local permutation tests (LPT) and conditional randomization tests (CRT) alternatives work and what statistic they use.

\subsubsection{How do we choose $Q_{Y|\X}$ in this set of experiments?} Given that our main objective in this experiment is comparing the power of different tests that use the same approximated distribution $Q_{Y|\X}$, we choose to fix $Q_{Y|X}=P^{(0)}_{Y|X}=\mathcal{N}(X,1)$, where $P^{(0)}_{Y|X}$ is the conditional distribution of $Y|X$ under $H_{0,\text{C2}}$.

\subsubsection{How do the LPT and CRT alternative tests work, and what statistics do they use?} We start explaining the two CRT alternative tests, which work in the same way but have different test statistics.
\begin{enumerate}
    \item We split our dataset $\mathcal{D}=\{(X_i,Y_i,Z_i\}_{i=1}^n$ in a training set $\mathcal{D}^{tr}=\{(X_i,Y_i,Z_i\}_{i=1}^{n^{tr}}$ and a test set $\mathcal{D}^{te}=\mathcal{D}-\mathcal{D}^{tr}$;
    
    \item We build an artificial training set $\widetilde{ \mathcal{D}^{tr}}=\{(X_i,\Tilde{Y}_i,Z_i)\}_{i=1}^{n^{tr}}$, where $\{\Tilde{Y}_i\}$ are sampled from $Q_{Y|\X}$;
    
    \item We train a probabilistic classifier $\hat{h}$ to distinguish samples from $\mathcal{D}^{tr}$ and $\widetilde{\mathcal{D}^{tr}}$, where the original set receives label 1 and the artificial data receives label 0;
    
    \item For $B \in \mathbb{N}$, we test each of the hypotheses of interest by computing a $p$-value of the form
    \begin{align*}
    p(\mathcal{D}^{te})=\frac{1+\sum_{j=1}^B \I\left[ T(\mathcal{D}^{te}) \leq T\left(\widetilde{ \mathcal{D}^{te}}^{(j)}\right)\right]}{B+1}, 
    \end{align*}
    where $T$ is a test statistic depending on $\hat{h}$ and each $\widetilde{\mathcal{D}^{te}}^{(j)}$ is obtained sampling different labels from $Q_{Y|\X}$. When $$T(\mathcal{D}^{te})=\frac{1}{|\mathcal{D}^{te}|}\sum_{(X,Y,Z) \in \mathcal{D}^{te}} \I[\hat{h}(X,Y,Z)>1/2]$$ we have CRT 1. When $$T(\mathcal{D}^{te})=\frac{1}{|\mathcal{D}^{te}|}\sum_{(X,Y,Z) \in \mathcal{D}^{te}} \hat{h}(X,Y,Z)$$ we have CRT 2.
\end{enumerate}

The LPT alternative tests work in the same way. The only difference is that the "artificial" variables $\{\Tilde{X_i}\}$ are obtained via local permutation within $Y$ levels.

$\hat{h}$ represents a CatBoost classifier in this experiment. When $\hat{h}$ is a logistic regressor (like in the KL tests), at least one test is too conservative.

\subsection{Digits experiment}

\begin{figure}[h!]
    \centering
    \hspace*{-.4cm}
    \begin{tabular}{cc}
    \includegraphics[width=.25\textwidth]{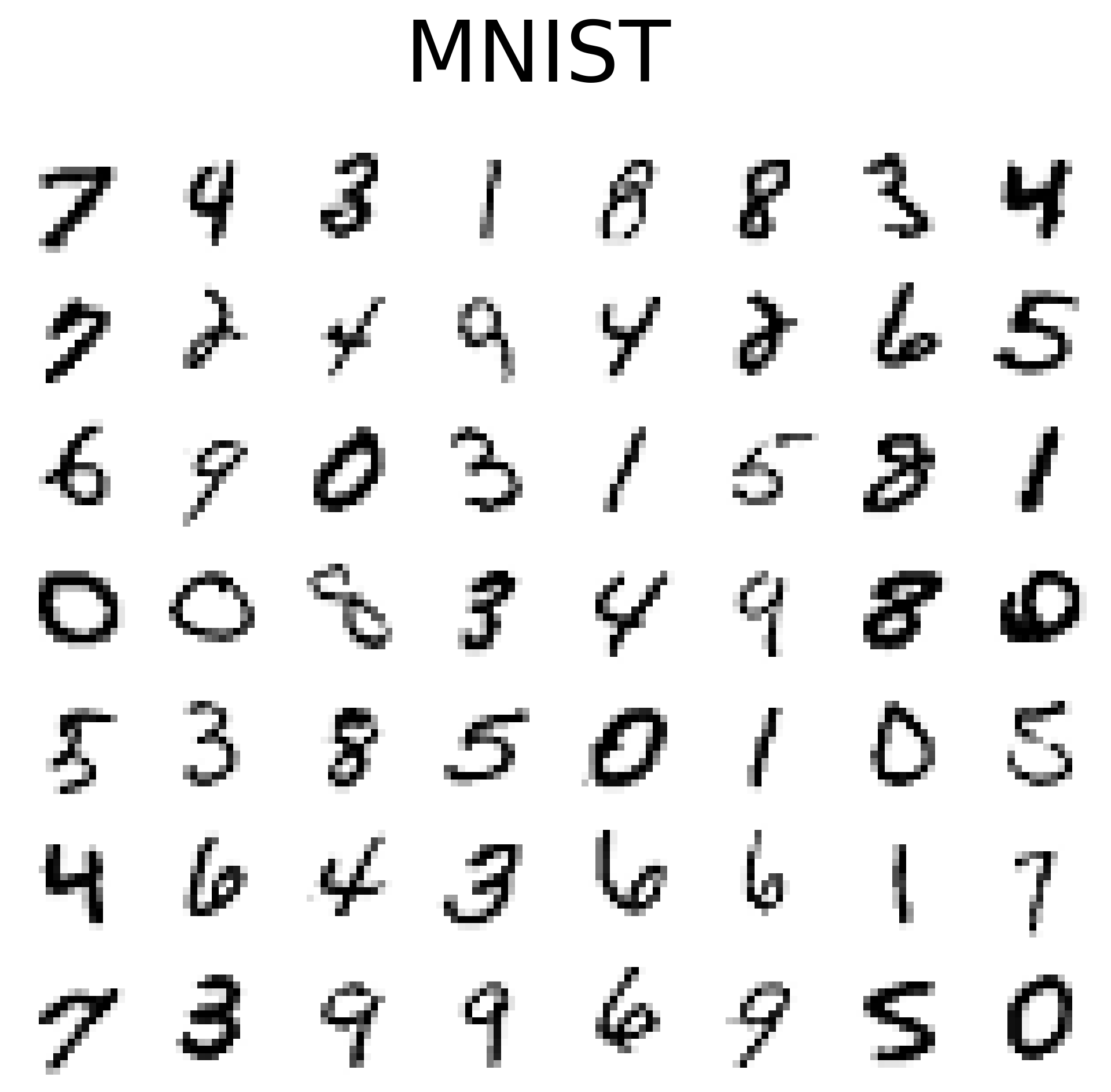}  & \includegraphics[width=.25\textwidth]{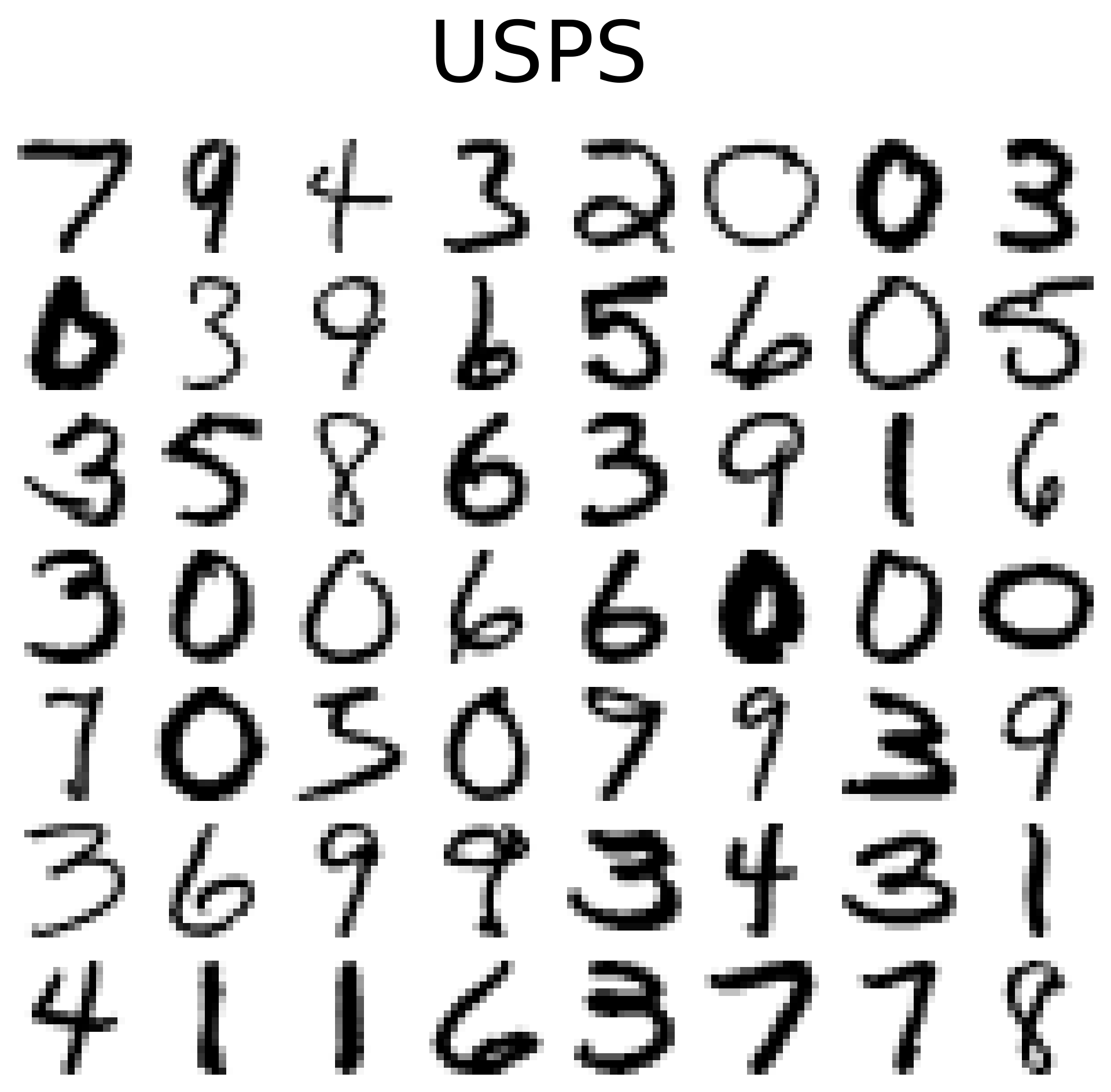}
    \end{tabular}
    \caption{Image samples from MNIST and USPS}
    \label{fig:digits_samples}
\end{figure}

MNIST samples tend to have more white pixels than USPS (Figure \ref{fig:digits_samples}); thus, the distributions of $\X$ are different in both datasets.

\subsection{A regression experiment}
\begin{figure}[h!]
    \centering
    \hspace*{-.4cm}
    \includegraphics[width=1\textwidth]{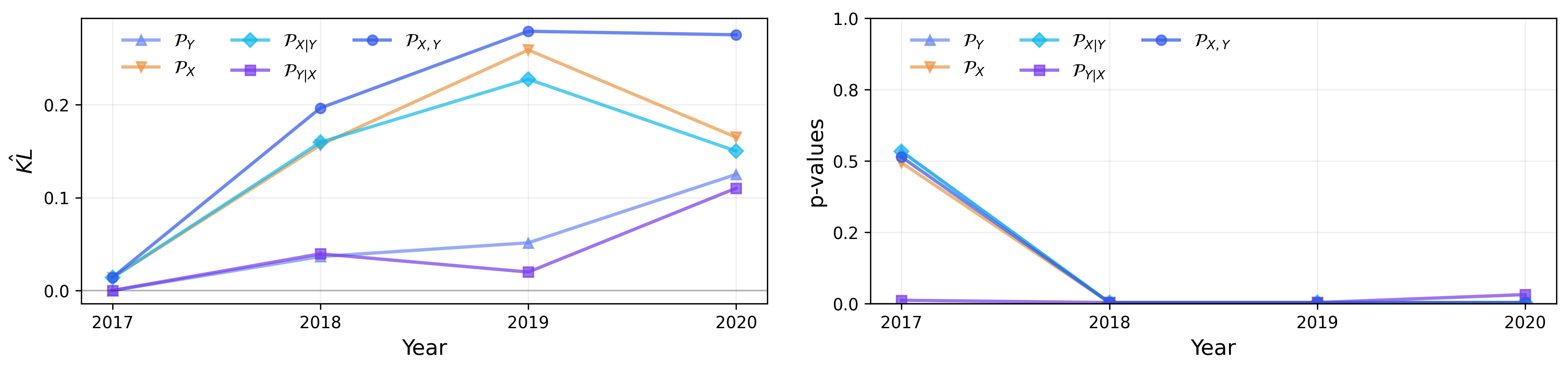}
    \caption{\small Detecting different types of dataset shift in a regression experiment. Using data from ENEM, the “Brazilian SAT,” we do a similar analysis compared to the credit one, comparing the probability distributions of 2018, 2019, and 2020 with the one in 2017. In this experiment, $Y$ is given by the student’s math score on the logarithmic scale, while $\X$ comprises students’ personal and socioeconomic features. It is possible to see that we detected all kinds of shifts in every moment after 2017. This result indicates that a model trained in 2017 might not generalize well to other years, and practitioners may consider retraining their models.} 
    \label{fig:enem}
\end{figure}

We present a regression experiment using data from 2017, 2018, 2019, and 2020 of ENEM\footnote{Data extracted from \url{https://www.gov.br/inep/pt-br/acesso-a-informacao/dados-abertos/microdados/enem}}, the “Brazilian SAT.” In each of the years, $Y$ is given by the students’ math score in a logarithmic scale while $\X$ is composed of six of their personal and socioeconomic features: gender, race, school type (private or public), mother’s education, family income and the presence of a computer at home. We randomly subsample the data in each one of the years to 30k data points with $10\%$ of them going to the test portion and then use the CatBoost algorithm to estimate the Radon-Nikodym derivative and the conditional distribution of $Y|\X$. When estimating the distribution of $Y|\X$, we first fit a regressor to predict $Y$ given $\X$, and then, using a holdout set, we fit a Gaussian model on the residuals. When testing for a shift in the $\X|Y$ distribution, we discretize $Y$ in 10 bins, evenly splitting the data. Even though we use the binned version of $Y$ to get the 
$p$-value, we report $\hat{\text{KL}}_{\X|Y}$ in the first panel of Figure \ref{fig:enem}. In this experiment, we do a similar analysis compared to the credit one, comparing the probability distributions of 2018, 2019, and 2020 with the one in 2017. From Figure \ref{fig:enem}, it is possible to see that we detected all kinds of shifts after 2017. This result indicates that a model trained in 2017 might not generalize well to other years, and practitioners may consider retraining their models from scratch using more recent data.

\newpage
\subsection{Detecting shifts with deep models}

\subsubsection{Extra plot}

\begin{figure*}[h]
    \centering
    \includegraphics[width=1\textwidth]{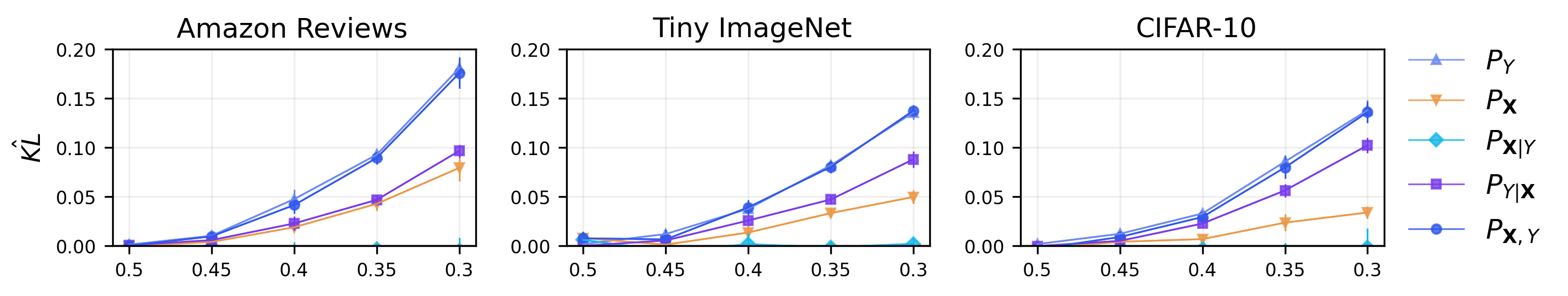}
    \caption{Detecting different types of dataset shift using complex high-dimensional data. For many values of $\delta$, we could detect all types of shift except conditional shift 1, which is expected because we introduced response shift in isolation ($\delta\neq.5$ is equivalent to label shift).}
\end{figure*}

\subsubsection{Extra results}

We include one extra experiment using the ``60k Stack Overflow Questions with Quality Rating" dataset\footnote{\url{https://www.kaggle.com/datasets/imoore/60k-stack-overflow-questions-with-quality-rate}} \cite{annamoradnejad2022multiview}. This dataset has three classes of rating from Stack Overflow questions. We repeat the same procedure to obtain results for the other deep-learning experiments.

\begin{figure}[H]
    \centering
    \begin{tabular}{c}
        ~~~ \begin{footnotesize}Ours\end{footnotesize} \\
        \includegraphics[width=.5\textwidth]{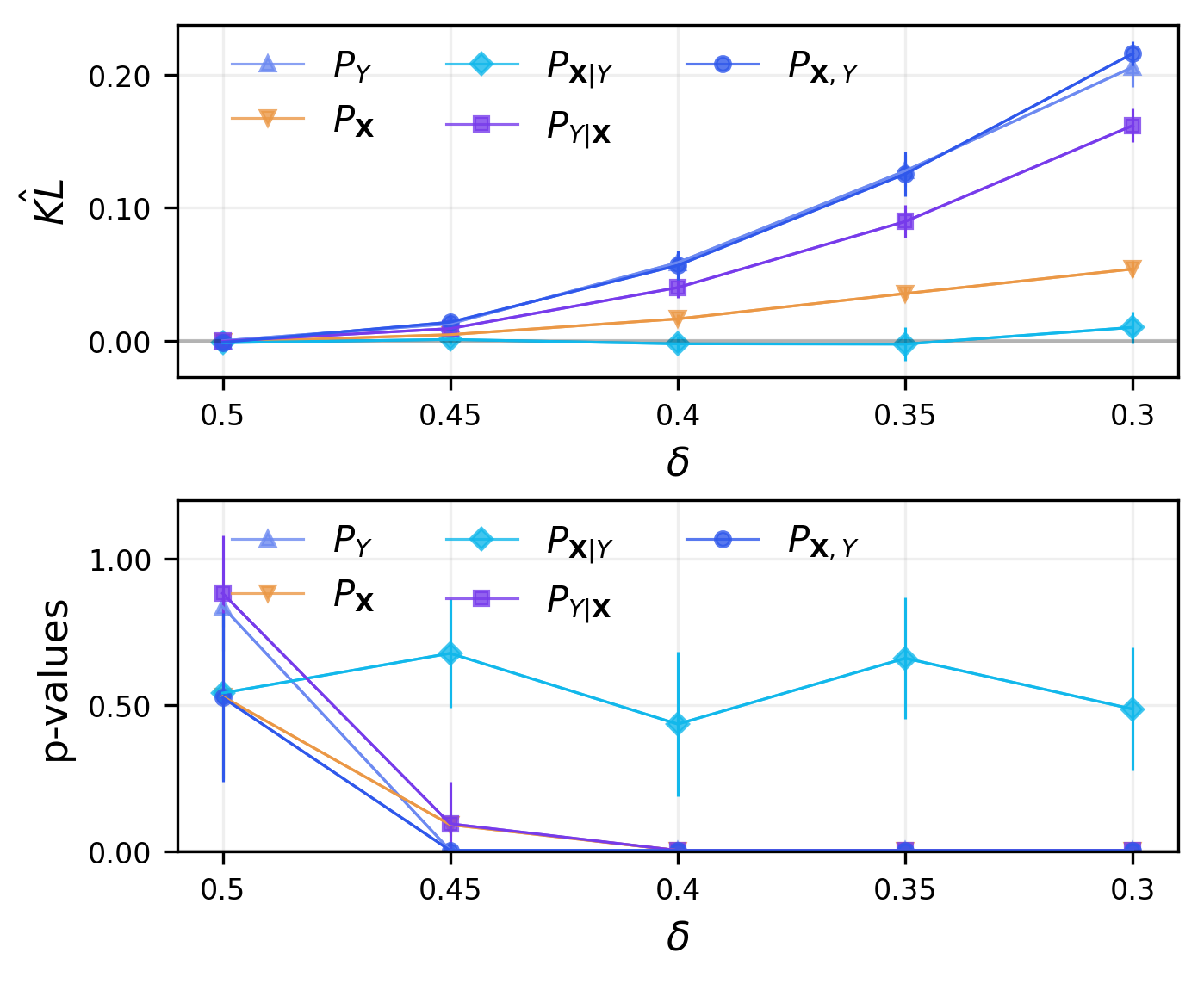}\\
        ~~~ \begin{footnotesize}Benchmark\end{footnotesize} \\ 
        \includegraphics[width=.5\textwidth]{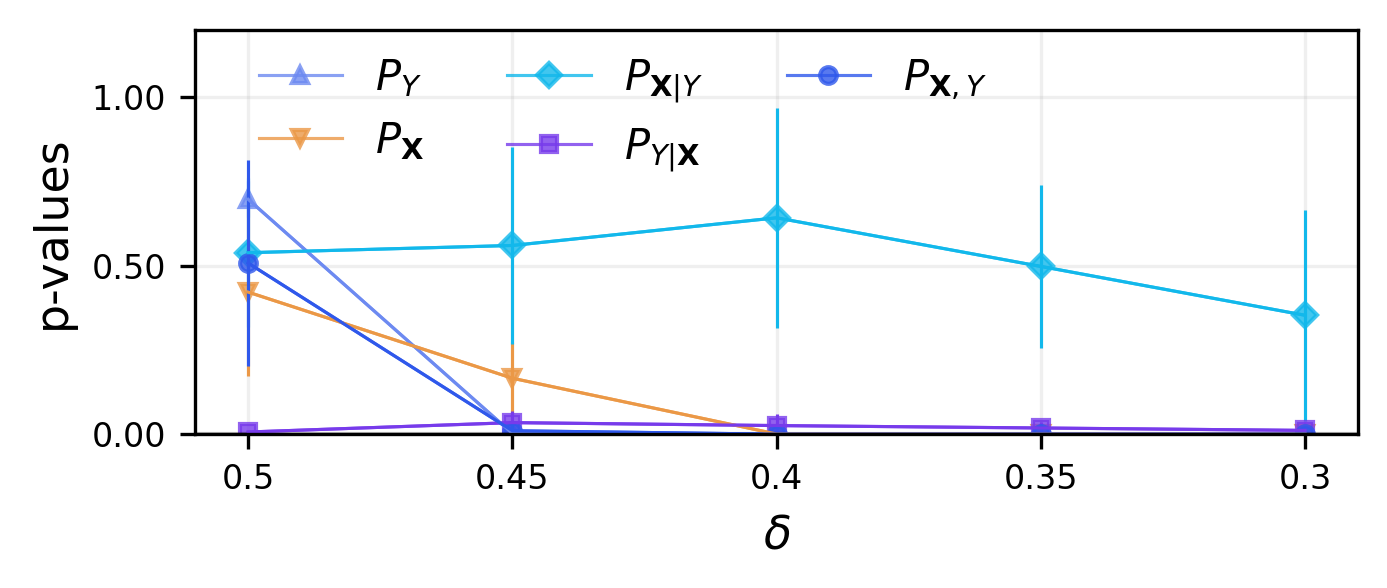}
    \end{tabular}
    \caption{\small Detecting different types of dataset shift using Stack Overflow text data. For many values of $\delta$, we could detect all types of shift except concept shift 1, which is expected because we introduced label shift in isolation ($\delta\neq.5$ is equivalent to label shift).}
    
\end{figure}

\subsection{Experiments running time}
All the experiments were run in a MacBook Air (M1, 2020) 16GB, except for the credit analysis experiment in an 80 CPUs Intel Xeon Gold 6148 cluster. We consider one iteration as all the steps needed to compute all the $p$-values used for a specific experiment. In the artificial data experiments, on average, each iteration performed by our framework took less than $1s$, while in real data experiments, each iteration took less than $200s$.